\documentclass[10pt,journal,compsoc]{IEEEtran}

\usepackage{amsmath,amsfonts}
\usepackage{algorithmic}
\usepackage{algorithm}

\usepackage{array}
\usepackage[caption=false,font=normalsize,labelfont=sf,textfont=sf]{subfig}
\usepackage{textcomp}
\usepackage{stfloats}
\usepackage{url}
\usepackage{verbatim}
\usepackage{graphicx}
\usepackage{multirow}
\usepackage{multicol}
\usepackage{threeparttable}
\usepackage{float}
\usepackage{stfloats}
\usepackage{marvosym}
\usepackage{xcolor}
\usepackage{etoolbox}
\usepackage{pifont}

\makeatletter
\patchcmd{\@makecaption}
  {\scshape}
  {}
  {}
  {}
\makeatother

\usepackage[pagebackref,breaklinks,colorlinks]{hyperref}
\usepackage[capitalize]{cleveref}
\crefname{section}{Sec.}{Secs.}
\Crefname{section}{Section}{Sections}
\Crefname{table}{Table}{Tables}
\crefname{table}{Tab.}{Tabs.}

\newcommand{\re}{\textcolor{black}}
\newcommand{\cmark}{\ding{51}}%
\newcommand{\xmark}{\ding{55}}%
\newcommand{\revise}{\textcolor{black}}

\begin{document}

\title{PDPP: Projected Diffusion for Procedure Planning in Instructional Videos}

\author{Hanlin Wang, Yilu Wu, Sheng Guo, \revise{Limin Wang,}~\IEEEmembership{Member,~IEEE}
\IEEEcompsocitemizethanks{
\IEEEcompsocthanksitem H. Wang, Y. Wu, and L. Wang are with the State Key Laboratory for Novel Software Technology, Nanjing University, Nanjing 210023, China. (E-mail: \{hlwang, yiluwu\}@smail.nju.edu.cn, lmwang@nju.edu.cn) 
\IEEEcompsocthanksitem L. Wang is also with Shanghai AI Laboratory, Shanghai 200232, China.
\IEEEcompsocthanksitem S. Guo is with MYbank, Ant Group, Hangzhou 310000, China. (E-mail: guosheng1001@gmail.com)
\IEEEcompsocthanksitem Corresponding author: L. Wang.
}  

\thanks{Manuscript received April 19, 2005; revised August 26, 2015.}}

\markboth{Journal of \LaTeX\ Class Files,~Vol.~14, No.~8, August~2015}%
{Shell \MakeLowercase{\textit{et al.}}: Bare Advanced Demo of IEEEtran.cls for IEEE Computer Society Journals}

\IEEEtitleabstractindextext{
\begin{abstract}
    In this paper, we study the problem of procedure planning in instructional videos, which aims to make a plan (i.e. a sequence of actions) given the current visual observation and the desired goal. 
    Previous works cast this as a sequence modeling problem and leverage either intermediate visual observations or language instructions as supervision \re{to make autoregressive planning}, resulting in complex learning schemes and expensive annotation costs.
    \re{To avoid intermediate supervision annotation and error accumulation caused by planning autoregressively, we propose a diffusion-based framework, coined as PDPP \revise{(Projected Diffusion model for Procedure Planning)}, to directly model the whole action sequence distribution with task label as supervision instead. Our core idea is to treat procedure planning as a distribution fitting problem under the given observations, thus transform the planning problem to a sampling process from this distribution during inference. The diffusion-based modeling approach also effectively addresses the uncertainty issue in procedure planning. Based on PDPP, we further apply joint training to our framework to generate plans with varying horizon lengths using a single model and reduce the number of training parameters required.
    We instantiate our PDPP with three popular diffusion models and investigate a series of condition-introducing methods in our framework, including condition embeddings, \revise{Mixture-of-Experts (MoEs)}, two-stage prediction and Classifier-Free Guidance strategy. Finally, we apply our PDPP to the \revise{Visual Planners for human Assistance (VPA)} problem which requires the goal specified in natural language rather than visual observation. We conduct experiments on challenging datasets of different scales and our PDPP model achieves the state-of-the-art performance on multiple metrics, even compared with those strongly-supervised counterparts. These results further demonstrates the effectiveness and generalization ability of our model.} Code and trained models are available at \url{https://github.com/MCG-NJU/PDPP}.
\end{abstract}

\begin{IEEEkeywords}
Goal-directed procedure planning, instructional video, diffusion model, conditional projection, AI assistants.
\end{IEEEkeywords}}

\maketitle

\IEEEdisplaynontitleabstractindextext

\IEEEpeerreviewmaketitle

\section{Introduction}
\label{sec::intro}
\begin{figure*}[t]
  \centering
  \includegraphics[width=1\linewidth]{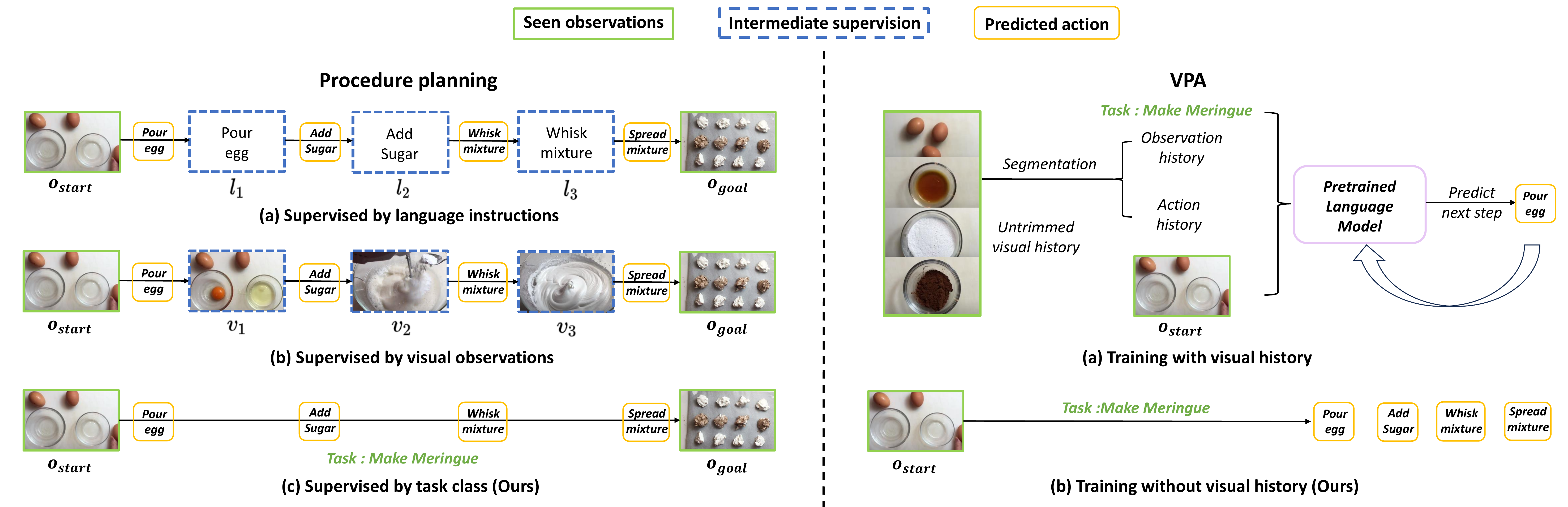}
   \caption{\re{Planning example.} \revise{"Seen observations" denotes the visual inputs for model, "Intermediate supervision" denotes additional supervision used in previous approaches, and "Predicted action" means the actions to be predicted by model.} For procedure planning, given a start observation $o_{start}$ and a goal state $o_{goal}$, the model is required to generate \re{an action sequence} that can transform $o_{start}$ to $o_{goal}$. Previous approaches rely on heavy intermediate supervision during training, while our model only needs the task class labels (shown in bottom row). \revise{We also evaluate another similar Visual Planners for human Assistance (VPA) problem, which takes visual inputs and a language-described goal to make plans.} We remove the requirement of visual history and plan directly with the given start observation and goal description.}
   \label{fig:question}
   \vspace{-10pt}
\end{figure*}

\IEEEPARstart{I}{nstructional} videos\cite{DBLP:conf/cvpr/ZhukovACFLS19,DBLP:conf/cvpr/AlayracBASLL16,DBLP:conf/cvpr/TangDRZZZL019} are strong knowledge carriers, which contain rich scene changes and various action steps. People can learn new skills from these videos by figuring out what actions should be performed to achieve the desired goals. Although this seems to be natural for humans, it is quite challenging for AI agents. Training a model that can learn how to make action plans from the start state to goal is crucial for the next-generation AI system as such a model is able to mine procedure knowledge from huge amounts of videos to  help people with goal-directed problems like cooking. Nowadays the computer vision community is paying growing attention to the instructional video understanding\cite{DBLP:conf/eccv/ChangHXAFN20, patel2023pretrained, DBLP:journals/tip/ZhaoRTZL22, DBLP:conf/cvpr/LiangWZY22, DBLP:conf/eccv/DvornikHPBMFJ22, DBLP:conf/nips/DvornikHDGJ21}. Among them, procedure planning in instructional vieos\cite{DBLP:conf/eccv/ChangHXAFN20} is becoming an important problem, which requires a model to produce goal-directed action plans given the current and goal visual observations. Different with traditional procedure planning problem in structured environments\cite{DBLP:conf/icra/FinnL17,DBLP:conf/icml/SrinivasJALF18}, this task needs to deal with unstructured environments and thus forces the model to learn structured and plannable representations from realistic videos. Specifically, given the visual observation at start and the desired goal, we need to produce a sequence of actions that transform the start state to the goal state, as shown in \cref{fig:question}.

Previous approaches for procedure planning in instructional videos often treat it as a sequence decision making problem and focus on predicting the individual action accurately in each step. 
\re{Since ground truth intermediate observations are not available for procedure planning, most autoregressive methods have to predict intermediate states and actions alternately with two-branch structure in a step-by-step manner\cite{DBLP:journals/ral/SunHLLZG22, DBLP:conf/eccv/ChangHXAFN20, DBLP:conf/iccv/BiLX21}.} Such models are complex and easy to accumulate errors during the planning process, especially for long sequences. 
Recently, Zhao \textit{et al.}\cite{DBLP:conf/cvpr/0004HDDWJ22} solves procedure planning with a single branch non-autoregressive model based on Transformer\cite{DBLP:conf/nips/VaswaniSPUJGKP17} to predict all intermediate steps in parallel. Yet, to obtain good performance, 
this method brings multiple learning objectives, complex training schemes, and tedious inference process. 
In contrast, we assume the procedure planning problem as a distribution fitting problem and thus \re{can be} solved with a sampling process. We directly model the joint distribution of the whole action sequence in instructional video rather than every discrete action. In this perspective, we can use a simple MSE loss to optimize our generative model and generate the whole action sequence at once, which contains less learning objectives and enjoys simpler training schemes.

For supervision in procedure planning, in addition to the action sequence, previous methods often need heavy intermediate visual\cite{DBLP:journals/ral/SunHLLZG22, DBLP:conf/eccv/ChangHXAFN20, DBLP:conf/iccv/BiLX21} or language\cite{DBLP:conf/cvpr/0004HDDWJ22} annotations for their learning process.
In contrast, we remove the requirement for these annotations and only use task labels from instructional videos as a condition for our model training (as shown in \cref{fig:question}), which could be easily obtained from the keyword or caption of the video with less labeling cost. Meanwhile, task information is closely related to the action sequences and can provide effective guidance for action planning. For example, in a video of $jacking$ $up$ $a$ $car$, the possibility of action $add \ sugar$ in it is almost zero. \revise{To some extent, supervision signals applied in previous methods are more specific task information. For instance, intermediate languages contain words unique to certain tasks, and intermediate observations include scenes relevant to specific tasks. These supervision signals implicitly contain the current task category, whereas the task labels we used explicitly leverage this information. We use the task-prediction model in the first stage to reduce the learning complexity of the later process. From the annotation perspective, task labels are much easier to obtain than intermediate supervision and are also easier to train with. Therefore, we chose to use task labels as the supervision signal.}

\revise{Previous approaches\cite{DBLP:journals/ral/SunHLLZG22, DBLP:conf/eccv/ChangHXAFN20, DBLP:conf/iccv/BiLX21, DBLP:conf/cvpr/0004HDDWJ22} for procedure planning train models with different horizons (i.e. the number of predicted action steps) separately, which means different models have to be applied when predicting with varied horizons. This setting presents some challenges as real-world tasks typically vary in the number of actions required to complete them. The main issue is that as the prediction horizon increases, the total number of trained parameters grows linearly, which is not conducive to development. We thus propose to train our model that can predict different planning horizons by joint training with data sequences of varied lengths. Our main goal is to reduce the number of training parameters to improve computation speed and avoid complex inference processes. To achieve this, embedding for horizon as condition and Mixture-of-Experts (MoEs)\cite{DBLP:conf/iclr/ShazeerMMDLHD17} are applied to our PDPP.}

In addition, modeling the uncertainty in procedure planning is a key factor for consideration. In reality, there might be more than one reasonable action sequences \re{that can} transform the given start state to goal state. For example, changing the order of $add$ $sugar$ and $add$ $butter$ in $making$ $cake$ process will not affect the final result. So action sequences can vary even with the same start and goal states. To address this problem, we consider adding randomness to our distribution-fitting process and perform training with a diffusion model\cite{DBLP:conf/nips/HoJA20,DBLP:conf/icml/NicholD21}. Solving procedure planning problem with a probabilistic diffusion model has two main benefits. First, a diffusion model changes the goal distribution to a random Gaussian noise by adding noise progressively to the initial data and learns the sampling process at inference time as an iterative denoising procedure starting from a random Gaussian noise. So randomness is naturally involved both for training and sampling in a diffusion model, which is helpful to model the uncertain action sequences for procedure planning. Second, it is easy to apply conditional diffusion process with the given start observation and goal state based on diffusion models, so we can model the procedure planning problem as a conditional sampling process with a simple training scheme. In this work, we directly concatenate conditions and action sequences together and propose a projected diffusion model to explicitly perform conditional diffusion process.

Finally, we extend our PDPP framework to a more challenging task called \revise{Visual Planners for human Assistance (VPA)}\cite{patel2023pretrained}. In this task, it argues that procedure planning is not useful in a real-world assistance setting due to the unavailability of the goal \re{observation}. Instead, VPA uses a language described goal like ``make a shelf". 
\re{So we can apply PDPP to the VPA problem by simply replacing the goal observation with task description}. This extension demonstrates the flexibility and generalization power of our PDPP framework.

In summary, our contributions are summarized as follows: 1) We cast the goal-directed procedure planning in instruction videos as a conditional distribution-fitting problem \re{and directly model the joint distribution of the whole action sequence}, which can be learned with a simple training scheme. 2) We introduce an efficient scheme for training the goal-directed planner, which avoids additional supervision of visual or language features and simply relies on task supervision instead. 3) We propose a projected diffusion model (PDPP) to learn the distribution of action sequences, which is able to naturally model the uncertainty in procedure planning. \re{Our PDPP is actually a multi-step generative prediction model based on probabilistic modeling rather than a simple action retrieval model. PDPP takes observation from the current environment and a video clip or text description as goal to make multi-step look ahead predictions and real-time inference. By introducing diffusion, our model is capable of generating multiple plausible action sequences and effectively modeling the uncertainty in planning task.}
4) \re{We apply joint training to PDPP to save the amount of training parameters. The resulting model can be more flexible and practical for procedure planning with varied horizons.} 5) Extensive experiments on three instructional videos datasets show that our method outperforms the existing state-of-the-art methods across different prediction horizons, demonstrating its generalization ability on goal-directed planning in instructional videos.

A preliminary version of this work\cite{wang2023pdppprojected} was published on the conference of IEEE CVPR 2023 as a highlight presentation. In this new version, we make the following extensions: \re{i) To save the amount of training parameters and make one single model handle planning with different horizons, we apply joint training to our model. We introduce horizon embedding and \revise{MoEs\cite{DBLP:conf/iclr/ShazeerMMDLHD17}} to PDPP to improve the model performance.} ii) We instantiate our PDPP with three popular diffusion model backbones: UNet\cite{DBLP:conf/miccai/RonnebergerFB15}, UNet with attention layers\cite{DBLP:conf/nips/HoJA20} and Transformer\cite{Peebles2022DiT}. For task supervision, we further divide it into three levels: no task supervision, task-level one-hot supervision, and task-level mask supervision. We study the performance of these backbones with different levels of task supervision on three datasets. iii) We propose to split the planning process into two stages to achieve better planning performance. That is, we first predict the first and last actions with the given start and goal observations. Then the predicted first and last actions are used as conditions to predict the intermediate actions. Experiments show that this two-stage planning process can bring better results on large scale dataset. \re{iv) We address the VPA problem by replacing the goal observation with task description in our PDPP framework, and our approach sets a new state-of-the-art performance on this task.} Experiments demonstrate that our methods have excellent generalization performance and can handle long-horizon prediction. We also provide more detailed analysis and ablations for the task supervision, joint training, and uncertainty modeling in our PDPP.

\section{Related work}
\label{sec::relatedwork}
\subsection{Procedural video understanding}
With an aim to learn the inter-relationship between different events or steps in videos, the problem of procedural video understanding, which is important for the training of AI models as human assistants, has gained more and more attention recently. Zhao \textit{et al.}\cite{DBLP:journals/tip/ZhaoRTZL22} investigated the problem of abductive visual reasoning, which requires vision systems to infer the most plausible visual explanation for the given visual observations. Furthermore, Liang \textit{et al.}\cite{DBLP:conf/cvpr/LiangWZY22} proposed a new task: given an incomplete set of visual events, AI agents are asked to generate descriptions not only for the visible events and but also for the lost events by logical inferring. Unlike these works trying to learn the abductive information of intermediate events, Chang \textit{et al.}\cite{DBLP:conf/eccv/ChangHXAFN20} introduced procedure planning in instructional videos which requires AI systems to plan an action sequence that can transform the given start observation to the goal state in order to assist humans in daily scenarios. Different with the long-term action anticipation (LTA) task\cite{DBLP:conf/cvpr/LohRF22, DBLP:conf/dagm/FarhaKSG20, DBLP:conf/wacv/MascaroAL23}, procedure planning requires a visual observation as goal to make goal-directed planning and covers a wide range of goal-oriented activities in everyday life. Patel \textit{et al.}\cite{patel2023pretrained} further point out the unavailability of the goal visual state in procedure planning and propose VPA task to use a language described goal instead. In this paper, we study the procedural video understanding problem by learning the goal-directed procedure planning task.

Multiple methods have been proposed to solve the procedure planning. Chang \textit{et al.}\cite{DBLP:conf/eccv/ChangHXAFN20} conduct a dual dynamic network to model the transformation relationship between intermediate states and actions. When inference, the actions and states are generated step by step. Bi \textit{et al.}\cite{DBLP:conf/iccv/BiLX21} follow this idea and use model-based reinforcement learning to predict the state-action pairs in an autoregressive way. They also learn the time-invariant context information from the given observations for better planning. Instead of using the intermediate states supervision, Zhao \textit{et al.}\cite{DBLP:conf/cvpr/0004HDDWJ22} apply a single branch non-autoregressive transformer decoder to predict all intermediate actions in parallel with language annotations. We here propose our PDPP model, which can generate both accurate and diverse plans without any states or language supervision, to solve this problem. 

\subsection{Diffusion probabilistic models} 
Diffusion probabilistic models\cite{DBLP:conf/icml/Sohl-DicksteinW15} are generative models which have been gaining significant popularity nowadays and have achieved great success in many areas. Ho \textit{et al.}\cite{DBLP:conf/nips/HoJA20} used a reweighted objective to train diffusion model and achieved great synthesis quality for image synthesis problem. Janner \textit{et al.}\cite{DBLP:conf/icml/JannerDTL22} studied the trajectory planning problem with diffusion model and get remarkable results. Besides, diffusion models are also used in video generation\cite{DBLP:journals/corr/abs-2204-03458, DBLP:journals/corr/abs-2210-02303}, density estimation\cite{DBLP:journals/corr/abs-2107-00630}, human motion\cite{DBLP:journals/corr/abs-2209-14916}, sound generation\cite{DBLP:journals/corr/abs-2207-09983}, text generation\cite{DBLP:journals/corr/abs-2205-14217} and other domains, all achieved competitive results. There are various architectures for implementing diffusion models, such as Unet\cite{DBLP:conf/miccai/RonnebergerFB15}, Unet with attention layers\cite{DBLP:conf/nips/HoJA20} and Transformer\cite{Peebles2022DiT}.

Diffusion Models work by destroying training data through the successive addition of Gaussian noise. Then these models learn to reverse this noising process by removing the added noises step by step to recover the initial data. Under such a design, diffusion models require no adversarial training and have the added benefits of learning scalability and parallelizability with a simple learning scheme. Besides, the generated results can vary with different noises, which is helpful to the diverse generation. With the above advantages of diffusion models, we in this work treat the whole action sequence in procedure planning as a distribution and apply diffusion process to this problem to model the uncertainty. Our proposed projected diffusion model PDPP achieves state-of-the-art performance only with a simple learning scheme.

\subsection{Projected gradient descent}
Projected gradient descent is an optimal solution suitable for constrained optimization problems, which is proven to be effective in optimization with rank constraints\cite{DBLP:journals/corr/ChenW15a}, online power system optimization problems\cite{DBLP:conf/allerton/HauswirthBHD16} and adversarial attack\cite{DBLP:conf/icip/DengK20}, etc. The core idea of projected gradient descent is adding a projection operation to the normal gradient descent method to project the output onto the feasible set, so that a function subject can be optimized to a constraint and the result is ensured to be in the feasible region. \re{In our PDPP, we apply diffusion to procedure planning to fit the goal distribution, which contains both the action sequence and condition-guided information. These conditions can be changed during the noise-adding and denoising stages. Inspired by projected gradient descent, we add a similar projection operation to our diffusion process, which keeps the conditional information for diffusion unchangeable and thus provides accurate guides for learning. The process of ``projecting back to a bounded normal range from an abnormal state'' is analogous to how we rectify a condition altered by noise to its accurate state, thereby remapping the overall distribution to a precisely guided range.}

\subsection{Mixture of Experts (MoEs)}
As an ensemble learning technology, \revise{MoEs\cite{DBLP:conf/iclr/ShazeerMMDLHD17}} divides a problem space into homogeneous regions by learning multiple expert modules to process the input signal, together with a gating module called as router to help combine the outputs
of different experts. When training, the router is learned to assign the combination weight across different experts, thus each expert is expected to focus on learning a certain sub-region of the input space. MoEs has shown its remarkable ability to scale neural networks such as LSTM\cite{DBLP:conf/iclr/ShazeerMMDLHD17}, CNN\cite{DBLP:journals/tip/AbbasA20, DBLP:conf/nips/YangBLN19} and Transformer\cite{DBLP:conf/nips/RiquelmePMNJPKH21, DBLP:conf/nips/ShazeerCPTVKHLH18, DBLP:conf/iclr/LepikhinLXCFHKS21}. By activating dynamic sub-networks for multiple inputs with routing strategy, MoEs provides separate parameters for different tasks and can thus reduce the bad influence of sharing parameters for conflicting inputs\cite{DBLP:conf/nips/ZhuZWWLWD22}. We in this paper apply MoEs to our Unet-attention based model on large scale dataset COIN\cite{DBLP:conf/cvpr/TangDRZZZL019} for better joint training. Specifically, we try two routing strategies: direct routing and learned routing for MoEs and apply this method to different parts of our model when planning with varied horizons.

\begin{figure*}[t]
  \centering
  \includegraphics[width=1\linewidth,height=0.42\linewidth]{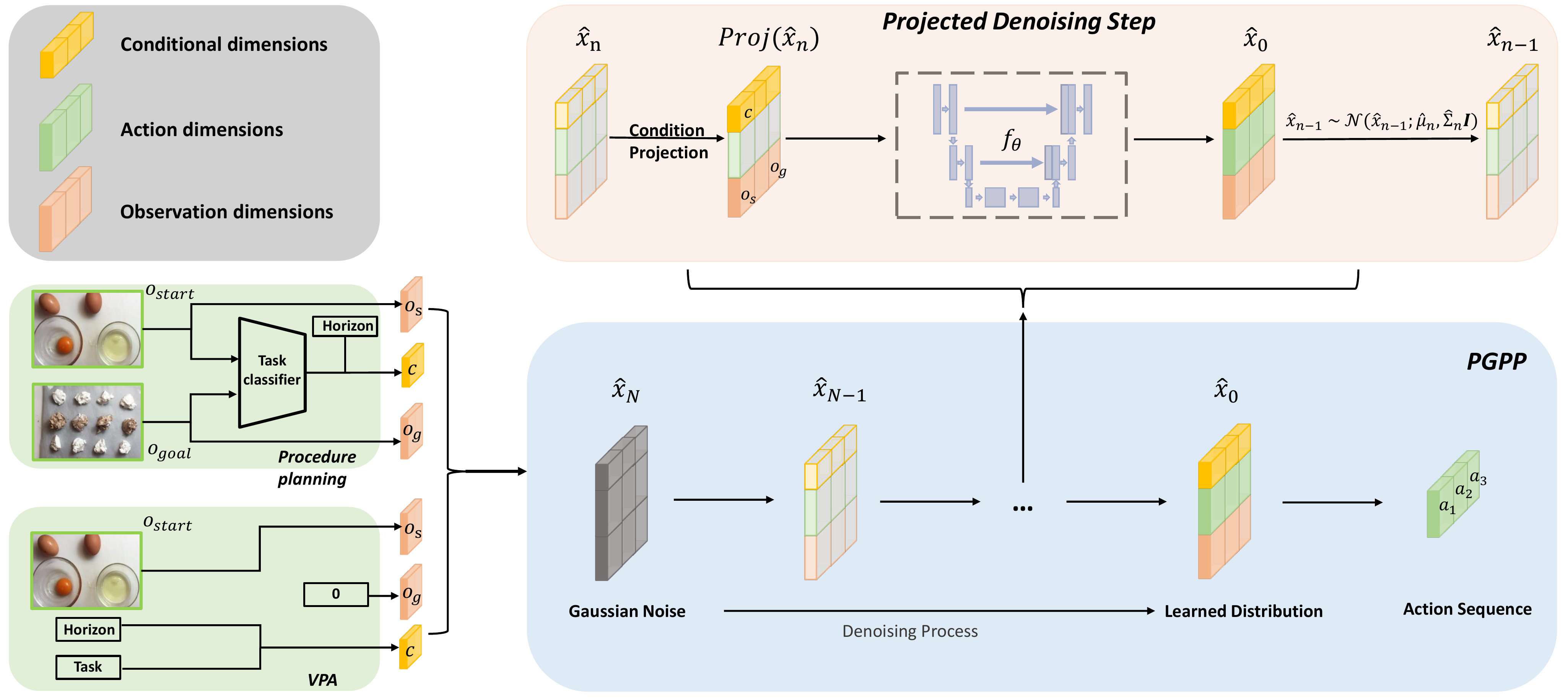}
   \caption{\re{Overview of PDPP (horizon $T$ = 3).} For procedure planning, we first train a task classifier to generate task condition, which will be used as guidance along with the given observations $o_s$, $o_g$ and horizon condition. Then we compute the denoising process iteratively. For VPA, task label is provided and $o_g$ is set as zero. In each step, we first conduct a condition projection to the input, then predict the initial distribution by the learned model $f_\theta$. After that we calculate $\hat{x}_{n-1}$ with the predicted $\hat{x}_{0}$. We finally select the action dimensions as our result after $N$ denoising steps.}
   \label{fig:overview}
\vspace{-3mm}
\end{figure*}

\section{Method}
\label{sec::method}
In this section, we present the details of our projected diffusion model for procedure planning in instructional videos (PDPP). An overview of PDPP is provided in \cref{fig:overview}. We will first introduce our setup for the planning problem in \cref{method:pro-formulation}. Then we present the diffusion model used to model the action sequence distribution and the way to add conditions in \cref{method:diffusion}. To provide more precise conditional guidance both for the training and sampling process, a simple projection method is applied to our model, which we will discuss in \cref{method:project}. \cref{method:a1aT} presents the two stage prediction process for procedure planning, which predicts \{$a_1, a_T$\} first and is proven to be useful for large scale dataset. Finally, we show the training scheme (\cref{method:train-scheme}) and sampling process (\cref{method:inference}) of our PDPP.

\subsection{Problem formulation}
\label{method:pro-formulation}

For procedure planning, we follow the problem set-up of Chang \textit{et al.}\cite{DBLP:conf/eccv/ChangHXAFN20}: given two video clips, start visual observation $o_{s}$ and visual goal $o_{g}$, a model is required to plan a sequence of actions $a_{1:T}$ so that the environment state can be transformed from $o_{s}$ to $o_{g}$. Here $T$ is the horizon of planning, which denotes the number of action steps for the model to take and \{$o_{s}$, $o_{g}$\} indicates two different environment states appeared in an instructional video. We decompose the procedure planning problem into two sub-problems, as shown in \cref{eq:1}. The first problem is to learn the task-related information $c$ with the given \{$o_{s}$, $o_{g}$\} pair. This can be seen as a preliminary inference for procedure planning. Then the second problem is to generate action sequences with the task-related information and given observations. Note that Bi \textit{et al.}\cite{DBLP:conf/iccv/BiLX21} also decompose the procedure planning problem into two sub-problems, but their purpose of the first sub-problem is to provide long-horizon information for the second stage since Bi \textit{et al.}\cite{DBLP:conf/iccv/BiLX21} plans actions step by step, while our purpose is to get condition for sampling to achieve an easier learning.

\begin{equation}
  \setlength{\abovedisplayskip}{-5pt}
  \setlength{\belowdisplayskip}{5pt}
  p(a_{1:T} | o_{s}, o_{g}) = \int p(a_{1:T} | o_{s}, o_{g}, c)p(c | o_{s}, o_{g}) dc.
  \label{eq:1}
\end{equation}

One of our motivations is to use one model for multiple planning horizons with joint training. That is, we sample one batch of action sequences for every prediction horizon and conducting gradient descent on all these data to complete one training step. Thus horizon information $h$ for every action sequence is also provided to our model as condition.

At training time, we first train a simple model (implemented as multi-layer perceptrons (MLPs)) with the given observations \{$o_{s}, o_{g}$\} to predict which the task category is. We use the task labels in instructional videos $\overline{c}$ to supervise the output $c$. After that, we evaluate $p(a_{1:T} | o_{s}, o_{g}, c, h)$ in parallel with our model and leverage the ground truth (GT) intermediate action labels as supervision for training. Compared with the visual and language supervision in previous works, task label supervision is easier to get and brings simpler learning schemes. At inference phase, we just use the start and goal observations to predict the task class information $c$ and then samples action sequences $a_{1:T}$ from the learned distribution with the given observations, horizon and predicted $c$, where $T$ is the planning horizon.

\subsection{Projected diffusion for procedure planning}
\label{method:diffusion}

As explained in \cref{method:pro-formulation}, the main part of our model is how to generate $a_{1:T}$ with given conditions to solve the procedure planning problem. Bi \textit{et al.}\cite{DBLP:conf/iccv/BiLX21} assume procedure planning as a Goal-conditioned Markov Decision Process and use a policy $p(a_t|o_t)$ along with a transition model $\tau_{\mu}(o_t|c,o_{t-1},a_{t-1})$ to perform the planning step by step, which is complex to train and slow for inference. We instead solve this as a direct distribution fitting problem with a diffusion model. In this section, we will first introduce the standard diffusion model. Then show our conditional sampling ways and our learning objective.

\begin{figure}
  \centering
  \includegraphics[width=1\linewidth]{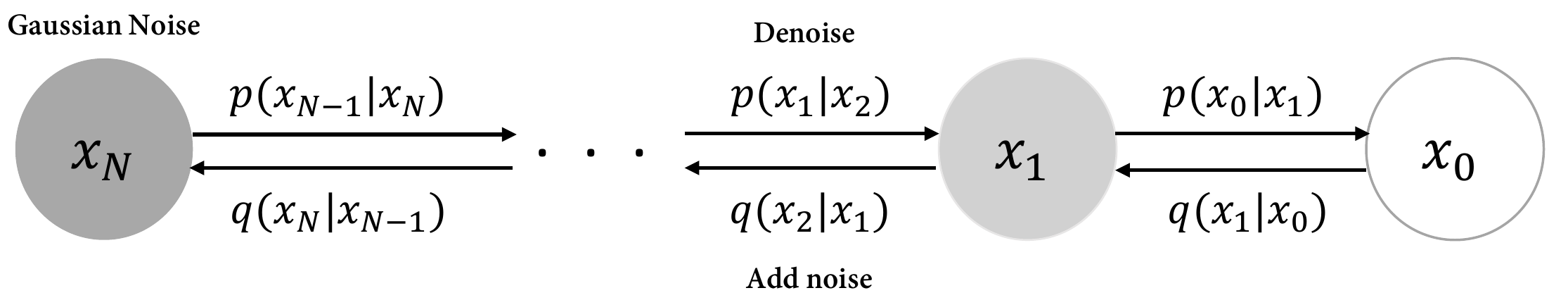}
   \caption{Schematic diagram for forward and reverse diffusion processes.}
   \label{fig:diff}
   \vspace{-10pt}
\end{figure}

~\\ \textbf{Diffusion model}. A diffusion model\cite{DBLP:conf/nips/HoJA20, DBLP:conf/icml/NicholD21} solves the data generation problem by modeling the data distribution $p(x_0)$ as a denoising Markov chain over variables \{$x_N...x_0$\} and assume $x_N$ to be a random Gaussian distribution. The forward process of a diffusion model is incrementally adding Gaussian noise to the initial data $x_0$ and can be represented as $q(x_n|x_{n-1})$, by which we can get all intermediate noisy latent variables $x_{1:N}$ with a diffusion step $N$. In the sampling stage, the diffusion model conducts iterative denoising procedure $p(x_{n-1}|x_n)$ for $N$ times to approximate samples from the target data distribution. The forward and reverse diffusion processes are shown in \cref{fig:diff}.

In a standard diffusion model, the ratio of Gaussian noise added to the data at diffusion step $n$ is pre-defined as $\{\beta_n \in (0, 1)\}_{n=1}^{N}$. Each adding noise step can be parametrized as

\begin{equation}
  \setlength{\abovedisplayskip}{-5pt}
  \setlength{\belowdisplayskip}{5pt}
  q(x_n|x_{n-1}) = \mathcal{N}(x_n; \sqrt{1-\beta_{n}} x_{n-1}, \beta_n\textbf{I}).
  \label{eq:forward}
\end{equation}
Since hyper-parameters $\{\beta_n\}_{n=1}^{N}$ are pre-defined, there is no training in the noise-adding process. As discussed in\cite{DBLP:conf/nips/HoJA20}, re-parameterize \cref{eq:forward} we can get:

\begin{equation}
  \setlength{\abovedisplayskip}{-5pt}
  \setlength{\belowdisplayskip}{5pt}
  x_n = \sqrt{\overline{\alpha}_n}x_0 + \sqrt{1 - \overline{\alpha}_n}\epsilon,
  \label{eq:forward_xt}
\end{equation}
where $\overline{\alpha}_n = \prod_{s=1}^{n}(1 - \beta_s)$ and $\epsilon \sim \mathcal{N}(0, I)$.

In the denoising process, each step is parametrized as:

\begin{equation}
  \setlength{\abovedisplayskip}{-8pt}
  \setlength{\belowdisplayskip}{5pt}
  p_\theta(x_{n-1}|x_n) = \mathcal{N}(x_{n-1};\mu_\theta(x_n,n),\Sigma_\theta(x_n,n)),
  \label{eq:sample}
\end{equation}
where $\mu_\theta$ is produced by a learnable model and $\Sigma_\theta$ can be directly calculated with $\{\beta_n\}_{n=1}^{N}$\cite{DBLP:conf/nips/HoJA20}. Then Ho \textit{et al.}\cite{DBLP:conf/nips/HoJA20} conduct a further derivation and set the learning objective for their diffusion model as the noise added to the uncorrupted data $x_0$ at each step. When training, the diffusion model first selects a diffusion step $n\in [1, N]$ and calculates $x_n$ as shown in \cref{eq:forward_xt}. Then the learnable model will compute $\epsilon_\theta(x_n, n)$ and calculate loss with the true noise add to the distribution at step $n$. After training, the diffusion model can simply generate data like $x_0$ by iteratively processing the denoising step starting from a random Gaussian noise.

However, such a diffusion model can not be applied to procedure planning problem since the sampling process in this task is condition-guided while no condition is applied in the standard diffusion model. There are multiple methods for implementing conditional-guided diffusion model, which could be divided into two categories: adding condition to diffusion model or conducting sampling guidance. The former adds conditional inputs to the model through cross-attention\cite{rombach2021highresolution}, AdaGN\cite{DBLP:conf/nips/DhariwalN21} or token concatenation along the sequence dimension\cite{DBLP:journals/corr/abs-2305-13311}, while the later conducts conditional sampling by providing explicit\cite{DBLP:conf/nips/DhariwalN21} or implicit\cite{DBLP:journals/corr/abs-2207-12598} guidance during the generation phase. Inspired by Janner \textit{et al.}\cite{DBLP:conf/icml/JannerDTL22}, we here choose to add condition to our diffusion model via designing conditional action sequence input and applying condition projection. Besides, we also tried to restrict predicted actions by task mask and apply MoEs to our model for task and horizon conditions.

Another noteworthy point is that the distribution we want to fit is the whole action sequence, which has a strong semantic information. We notice that in experiments which take noise as the predicting objective like\cite{DBLP:conf/nips/HoJA20}, our model just fails to be trained, indicating that directly predicting the semantically meaningless noise sampled from random Gaussian distribution can be hard. To address this problem, we change the learning objective of our model to provide a clear and strong learning distribution.

~\\ \textbf{Conditional action sequence input}. The input of a standard diffusion model is the data distribution it needs to fit and no guided information is required. For the procedure planning problem, the distribution we aim to fit is the intermediate action sequences $[a_1, a_2...a_T]$, which depends on the given observations. Thus we need to find how to add these guided conditions into the diffusion process. For a concise learning strategy, we here apply a simple way to achieve our goal: just treat these conditions as additional information and concatenate them along the action feature dimension. Specifically, our conditions include observation(o), task label(c) and horizon(h) for joint learning. For observation, since the start/end observations are more related to the first/last actions, we concatenate $o_s$ and $a_1$ together, same for $o_g$ and $a_T$. Task label and horizon features are useful to the whole predicted sequence, thus we duplicate them for every action. Thus our model input for training now can be represented as a multi-dimension array:

\begin{equation}
  \setlength{\abovedisplayskip}{-8pt}
  \setlength{\belowdisplayskip}{0pt}
    \begin{bmatrix}
        h&  h&  & h &h\\
        c&  c&  & c &c\\
        a_1&  a_2&  ...& a_{T-1} & a_T\\
        o_s&  0&  & 0 & o_g
    \end{bmatrix}.
    \label{input}
\end{equation}

Each column in our model input represents a certain action and the corresponding condition information. Note that we do not need the intermediate visual observations as supervision, so all observation dimensions are set to zero except for the start and end observations. For VPA, $o_g$ is not provided and thus set as zero.

~\\ \textbf{Adding condition by task mask and MoEs}. Note that the conditional action sequence input can change with different ways to add task and horizon conditions. That is, task label dimension is removed when task mask condition is applied and horizon dimension will be discarded when using MoEs.

Task mask can be seen as a more adequate application of task information. Once the task category is determined, the planning sequence can only consist of actions belonging to this task. For example, actions in task "Build Simple Floating Shelves" are "cut shelve", "assemble shelve", "sand shelve", "paint shelve" and "attach shelve". Only these five actions can be involved in any action sequence describing how to build simple floating shelves. So with the predicted or given task condition, we can get a task mask which blocks out all actions that are not part of this task and thus reduce the searching space for action planning. In this sense, we can introduce task condition by multiplying the action distribution by the task mask.

As for horizon condition, in addition to concatenating horizon information and action together, we also apply \revise{MoEs\cite{DBLP:conf/iclr/ShazeerMMDLHD17}}. \revise{The main idea of MoEs is to replicate certain parameters within a model and use routing algorithms to process different inputs with different sets of parameters. Thus, we replicate the parts of our model where MoEs is used and provide condition information to the routing gate module added in PDPP. We implement both direct routing and learned routing strategies. Direct routing just distributes input to specified sub-networks according to the horizon condition while learned routing learns a weight to integrate the outputs from all sub-networks together with the given conditions. Specifically, the learned routing module is implemented as MLP. Since our main goal is to save model parameters by joint training, we here only consider applying MoEs to the convolution part or attention part of Unet with attention layers model.
By introducing MoEs, we hope PDPP can learn how to handle inputs with different planning horizons to minimize the negative impact brought by joint training.}

~\\ \textbf{Learning objective of our diffusion model}. As mentioned above, the learning objective of a standard diffusion model is the random noise added to the distribution at each diffusion step. This learning scheme has demonstrated great success in visual data synthesis area. For procedure planning, however, the distribution we need to fit contains high-level features rather than raw pixels. Predicting random noise for procedure planning results in fault training, which is similar to text generation\cite{DBLP:journals/corr/abs-2205-14217}. One possible reason is that the action sequence distribution we need to predict has strong semantics. When the noise predicted in the early stage of sampling is not very accurate, the feature distribution after initial denoising steps can not have the required strong semantics, which leads to the following denoising operation deviates from correct direction and get meaningless sampling result. So we modify the learning objective to the initial input $x_0$, which is described in \cref{method:train-scheme}.

\vspace{-5pt}
\subsection{Condition projection during learning}
\label{method:project}

Our model transforms a random Gaussian noise to the final result, which has the same structure with our model input by conducting $N$ denoising steps. Since we combine the conditional information with action sequences as the data distribution, these conditional guides can be changed during the denoising process. However, the change of these conditions will bring wrong guidance for the learning process and make the conditions useless. To address this problem, we add a condition projection operation into the learning process. That is, we force the condition dimensions not changed during training and inference by assigning the initial value. The input $x$ of condition projection is either a noise-add data (Alg.1 L5) or the predicted result of model (Alg.1 L7). We use $\{\hat{c}, \hat{a}, \hat{o}, \hat{h}\}$ to represent different dimensions in $x$, then our projection operation $\mathrm{Proj}()$ can be denoted as:

\begin{equation}
  \setlength{\abovedisplayskip}{-5pt}
  \setlength{\belowdisplayskip}{5pt}
\begin{aligned}
    \begin{bmatrix}
        \hat{h}_1&  \hat{h}_2&  &  &\hat{h}_T\\
        \hat{c}_1&  \hat{c}_2&  &  &\hat{c}_T\\
        \hat{a}_1&  \hat{a}_2&  ...&  & \hat{a}_T\\
        \hat{o}_1&  \hat{o}_2&  &  & \hat{o}_T
    \end{bmatrix}
    \to
    \begin{bmatrix}
        h&  h&  &  &h\\
        c&  c&  &  &c\\
        \hat{a}_1&  \hat{a}_2&  ...&  & \hat{a}_T\\
        o_s&  0&  &  & o_g
    \end{bmatrix},
    \\x \qquad \qquad \qquad \qquad \mathrm{Proj}(x) \quad \qquad
  \label{eq:project}
\end{aligned}
\end{equation}
where $\hat{h}_i$, $\hat{c}_i$, $\hat{o}_i$ and $\hat{a}_i$ denote the $i^{th}$ horizon, task, observation dimensions and predicted action logits in $x$, respectively. $h$, $c$, $o_s,o_g$ are the conditions. When task mask is applied, condition projection for task information is replaced with multiplying the action distribution by the task mask. For joint training with MoEs, horizon condition is used in routing gate module and there is no condition projection for horizon dimensions.

\vspace{-5pt}
\subsection{Two stage prediction process}
\label{method:a1aT}

On the basis of discussions in \cref{method:pro-formulation}, we further split $p(a_{1:T} | o_{s}, o_{g}, c, h)$ modeling process into two stages for procedure planning. Considering that the given observations are visual states around the start and end actions, we propose to predict $a_1, a_T$ first, then the predicted two actions can be regarded as action conditions for planning and stay unchanged with condition projection. The two stage prediction process makes full use of the given observations and provides more guidance to the sampling process. One problem with this scheme is that the action condition used in training is the ground truth, but $a_1, a_T$ predicted by model cannot be completely correct, thus the distribution difference between the training set and test set may be further amplified, resulting in overfitting problem. Experiments show that this strategy can benefit our model on large scale dataset, while performs bad for datasets with limited size, which confirms our thought.

\begin{algorithm}[bp]
    \algsetup{linenosize=\small} \small
    \caption{Training}
    \begin{algorithmic}[1]
        \REQUIRE Initial input $x_0$, total diffusion steps number $N$, model $f_\theta$, $\{\overline{\alpha}_n\}_{n=1}^{N}$, weight matrix $w$
        \REPEAT
        \STATE $n \sim Uniform(\{1,...,N\})$
        \STATE $\epsilon \sim \mathcal{N}(0, I)$
        \STATE $x_n = \sqrt{\overline{\alpha}_n}x_0 + \sqrt{1 - \overline{\alpha}_n}\epsilon$
        \STATE $\hat{x}_0 = f_\theta(Proj(x_n), n)$
        \STATE Take gradient descent step on
        \STATE \quad $\bigtriangledown_\theta \left |\left | \left (x_0 - Proj(\hat{x}_0) \right ) * w \right | \right |^2$
        \UNTIL converged
        
    \end{algorithmic}
    \label{train_alg}
\end{algorithm}

\section{Training And Inference}
\subsection{Training scheme}
\label{method:train-scheme}

Our training scheme contains two stages: a) training a task-classifier model $\mathcal{T}_\phi (c  | o_s, o_g)$ that extracts conditional guidance from the given start and goal observations; b) leveraging the projected diffusion model to fit the target action sequence distribution. 

For the first stage, we apply MLP models to predict the task class $c$ with the given $o_s,o_g$. Ground truth task class labels $\overline{c}$ are used as supervision. In the second learning stage, we follow the basic training scheme for diffusion model, but change the learning objective as the initial input $x_0$. We denote our learnable model as $f_\theta(x_n, n)$ and our training loss is:

\begin{equation}
  \setlength{\abovedisplayskip}{-10pt}
  \setlength{\belowdisplayskip}{5pt}
  \mathcal{L}_{\mathrm{diff}} = \sum_{n=1}^{N}(f_\theta(x_n, n) - x_0)^2 ,
  \label{eq:task_loss}
\end{equation}

We believe that predicting actions corresponding to the given observations are more important because they can be inferred directly from observation conditions. Thus we rewrite \cref{eq:task_loss} as a weighted loss by assigning a greater weight to \{$a_1, a_T$\} in procedure planning 
with a weight matrix 
 $w$.

Besides, we add a condition projection step to our diffusion process. So given the initial input $x_0$ which contains action sequences, task conditions and observations, we first add noise to the input to get $x_n$, and then apply condition projection to ensure the guidance not changed. With $x_n$ and the corresponding diffusion step $n$, we calculate the denosing output $f_\theta(x_n, n)$, followed by condition projection again. Finally, we compute the weighted $\mathcal{L}_{\mathrm{diff}}$ and update model, as shown in \cref{train_alg}. Note that the planning horizon of $x_0$ can vary for joint training.

\subsection{Inference}
\label{method:inference}

At inference time, only the start observation $o_s$ and goal observation $o_g$ are provided for procedure planning. We thus first predict the task class by choosing the maximum probability value in the output of task-classifier model $\mathcal{T}_\phi$. Then the predicted task class $c$ is used as the task condition. 
To sample from the learned action sequence distribution, we start with a Gaussian noise, and iteratively conduct denoise and condition projection for $N$ times. The detailed inference process is shown in \cref{infer_alg}.

\begin{algorithm}
    \algsetup{linenosize=\small} \small
    \caption{Inference}
    \begin{algorithmic}[1]
        \REQUIRE total diffusion steps number $N$, model $f_\theta$, $\{\overline{\alpha}_n\}_{n=1}^{N}$, $\{\beta_n\}_{n=1}^{N}$
        
        \STATE $\hat{x}_N\sim \mathcal{N}(0,I)$
        \FOR{$n=N,...,1$}
            \STATE $\hat{x}_0 = f_\theta(Proj(\hat{x}_n), n)$
            \IF{$n > 1$} 
            \STATE $\hat{\mu}_n = \frac{\sqrt{\overline{\alpha}_{n-1}}\beta_n}{1-\overline{\alpha}_n} \hat{x}_0 + 
            \frac{\sqrt{\alpha_n}(1-\overline{\alpha}_{n-1})}{1-\overline{\alpha}_n} \hat{x}_n$
            \STATE $\hat{\Sigma}_n = \frac{1-\overline{\alpha}_{n-1}}{1-\overline{\alpha}_n} \cdot \beta_n$
            \STATE $\hat{x}_{n-1}\sim \mathcal{N}(\hat{x}_{n-1};\hat{\mu}_n,\hat{\Sigma}_n\textbf{I})$
            \ENDIF
        \ENDFOR
        \STATE return $\hat{x}_0$
    \end{algorithmic}
    \label{infer_alg}
\end{algorithm}

To further accelerate the sampling process, we apply DDIM\cite{DBLP:conf/iclr/SongME21} to our model and thus generate output with $N'(< N)$ sampling steps. The sampling process with DDIM is presented in \cref{infer_alg_ddim}.

\begin{algorithm}
    \algsetup{linenosize=\small} \small
    \caption{Inference with DDIM}
    \begin{algorithmic}[1]
        \REQUIRE DDIM sampling timesteps $[t_1, t_2, ..., t_{N'}]$, model $f_\theta$, $\{\overline{\alpha}_{t_n}\}_{n=1}^{N'}$, hyperparameter $\eta$
        
        \STATE $\hat{x}_{t_{N'}}\sim \mathcal{N}(0,I)$
        \FOR{$n=N',...,1$}
            \STATE $\hat{x}_0 = f_\theta(Proj(\hat{x}_{t_n}), t_n)$
            \STATE $\sigma_{t_n} = \eta \sqrt{(1-\overline{\alpha}_{t_{n-1}})/(1 -\overline{\alpha}_{t_n})} \cdot \sqrt{1 - \overline{\alpha}_{t_n} / \overline{\alpha}_{t_{n-1}}}$
            
            \IF{$n > 1$} 
            \STATE $\hat{\mu}_{t_n} = \sqrt{\overline{\alpha}_{t_{n-1}}}\hat{x}_0 + \sqrt{1-\overline{\alpha}_{t_{n-1}} - \sigma_{t_n}^2} \cdot \frac{\hat{x}_{t_n} - \sqrt{\overline{\alpha}_{t_n}}\hat{x}_0}{\sqrt{1 - \overline{\alpha}_{t_n}}}$
            \STATE $\hat{x}_{t_{n-1}}\sim \mathcal{N}(\hat{x}_{t_{n-1}};\hat{\mu}_{t_n},\sigma_{t_n}^2\textbf{I})$
            \ENDIF
        \ENDFOR
        \STATE return $\hat{x}_0$
    \end{algorithmic}
    \label{infer_alg_ddim}
\end{algorithm}

Once we get the predicted output $\hat{x}_0$, we take out the action sequence dimensions $[\hat{a}_1,...,\hat{a}_T]$  and select the index of every maximum value in $\hat{a}_i (i=1,...,T)$ as the action sequence plan. Note that in the training stage of procedure planning, the class condition dimensions of $x_0$ are the ground truth task labels, not the output of our task-classifier as in inference.

\section{Experiments}

In this section, we evaluate our PDPP model on three real-life datasets and show our competitive results. We first present the result of our first training stage, which predicts the task class with the given observations in \cref{exp:task}. Then we conduct detailed ablation studies to provide a thorough analysis on our proposed PDPP in \cref{exp:ablation}. After that, we compare our performance with other alternative approaches on the three datasets and demonstrate the effectiveness of our model in \cref{exp:compare}. \re{For a more comprehensive evaluation, we analyze the failure cases and discuss the computational efficiency of PDPP in \cref{exp:error} and \cref{exp:efficiency}, respectively.} Finally, we show our prediction uncertainty evaluation results in \cref{exp:uncertainty}. For VPA, we train our PDPP with best settings obtained from procedure planning and the results are presented in \cref{exp:VPA}. To avoid results influenced by initial random noise when sampling, we calculate the mean values of multiple sampling results with different initial random noises as our results. Unless otherwise specified, we use the DDIM sampling process to get all results.

\vspace{-5pt}
\subsection{Evaluation protocol}
\label{exp:protocol}

\vspace{3pt}
\noindent \textbf{Datasets.} We evaluate our PDPP model on three instructional video datasets: \textbf{CrossTask}\cite{DBLP:conf/cvpr/ZhukovACFLS19}, \textbf{NIV}\cite{DBLP:conf/cvpr/AlayracBASLL16}, and \textbf{COIN}\cite{DBLP:conf/cvpr/TangDRZZZL019}. CrossTask contains 2,750 videos from 18 different tasks, with an average of 7.6 actions per video. The NIV dataset consists of 150 videos about 5 daily tasks, which has 9.5 actions in one video on average. COIN is much larger with 11,827 videos, 180 different tasks and 3.6 actions/video.

For procedure planning, We randomly select 70\% data for training and 30\% for testing as previous work\cite{DBLP:conf/eccv/ChangHXAFN20,DBLP:conf/iccv/BiLX21,DBLP:conf/cvpr/0004HDDWJ22}.  
Following previous work\cite{DBLP:conf/eccv/ChangHXAFN20,DBLP:conf/iccv/BiLX21,DBLP:conf/cvpr/0004HDDWJ22}, we extract all action sequences $\{[a_i, ..., a_{i+T-1}]\}_{i=1}^{n-T+1}$ with predicting horizon $T$ from the given video which contains $n$ actions by sliding a window of size $T$. Then for each action sequence $[a_i, ..., a_{i+T-1}]$, we choose the video clip feature at the beginning time of action $a_i$ and clip feature around the end time of $a_{i+T-1}$ as the start observation $o_s$ and goal state $o_g$, respectively. Both clips are 3 seconds long. For experiments conduct on CrossTask, we use two kinds of pre-extracted video features as the start and goal observations. One are the features provided in CrossTask dataset: each second of video content is encoded into a 3200-dimensional feature vector as a concatenation of the I3D, ResNet-152 and audio VGG features\cite{DBLP:conf/cvpr/HeZRS16,DBLP:conf/icassp/HersheyCEGJMPPS17,DBLP:conf/cvpr/CarreiraZ17}, which are also applied in\cite{DBLP:conf/iccv/BiLX21,DBLP:conf/eccv/ChangHXAFN20}. The other kind of features are generated by the encoder trained with the HowTo100M\cite{DBLP:conf/iccv/MiechZATLS19} dataset, as in\cite{DBLP:conf/cvpr/0004HDDWJ22}. For experiments on the other two datasets, we follow\cite{DBLP:conf/cvpr/0004HDDWJ22} to use the HowTo100M features for a fair comparison.

For VPA, we follow Patel \textit{et al.}\cite{patel2023pretrained} to conduct our experiments on CrossTask and COIN. Ratios of training set and test set are 7/3 and 9/1 for these two datasets, respectively. Action sequences are extracted as in procedure planning, and the start observation video clip is 4 seconds long. All video features are get with the encoder trained on HowTo100M dataset. Note that our PDPP do not require video history as additional input.

~\\ \noindent \textbf{Metrics}. Following previous work\cite{DBLP:conf/eccv/ChangHXAFN20,DBLP:conf/iccv/BiLX21,DBLP:conf/cvpr/0004HDDWJ22}, we apply three metrics to evaluate the performance. a) \textbf{Success Rate (SR)} considers a plan as a success only if every action matches the ground truth sequence. b) \textbf{mean Accuracy (mAcc)} calculates the average correctness of actions at each individual time step, which means an predicted action is considered correct if it matches the action in ground truth at the same time step. c) \textbf{mean Intersection over Union (mIoU)} measures the overlap between predicted actions and ground truth by computing Iou $\frac{|\{a_t\} \cap \{\hat{a_t}\}|}{|\{a_t\} \cup \{\hat{a_t}\}|}$, where $\{a_t\}$ is the set of ground truth actions and $\{\hat{a_t}\}$ is the set of predicted actions. Previous approaches\cite{DBLP:conf/cvpr/0004HDDWJ22,DBLP:conf/iccv/BiLX21} compute the mIoU metric on every mini-batch (batch size larger than one) and calculate the average as the result. This brings a problem that the mIoU value can be influenced heavily by batch size. Consider if we set batch size is equal to the size of training data, then all predicted actions can be involved in the ground truth set and thus be correct predictions. However, if batch size is set to one, then any predicted action that not appears in the corresponding ground truth action sequence will be wrong. To address this problem, we standardize the way to get mIoU as computing IoU on every single sequence and calculating the average of these IoUs as the result (equal to setting of batch size $= 1$).

\vspace{-10pt}
\revise{\subsection{Implementation details}}
\label{method:implementation}
\vspace{-10pt}
\revise{~\\ \textbf{Architectures}.}
\revise{We use three backbones to implement our learnable model for projection diffusion. }

\revise{~\\ $\bullet$ \textit{Unet}\cite{DBLP:conf/miccai/RonnebergerFB15}. We implement our model as a basic 3-layer Unet. As in\cite{DBLP:conf/icml/JannerDTL22}, each layer in our model consists of two residual blocks\cite{DBLP:conf/cvpr/HeZRS16} and one downsample or upsample operation. One residual block consists of two convolutions, each followed by a group norm\cite{DBLP:conf/eccv/WuH18} and Mish activation function\cite{DBLP:journals/corr/abs-1908-08681}. Time embedding is produced by a fully-connected layer and added to the output of first convolution. We apply a 1d-convolution along the planning horizon dimension as the downsample/upsample operation. We set the kernel size of 1d-convolution as $2$, stride as $1$, padding as $0$ so the length change of planning horizon dimension keeps $1$ after each downsample or upsample. The hidden dimensions for each layer are 256, 512, 1024, respectively.}
\revise{~\\ $\bullet$ \textit{Unet with attention layers}\cite{DBLP:conf/nips/HoJA20}: We implement our Unet-attention architecture by adding self-attention\cite{DBLP:conf/nips/VaswaniSPUJGKP17} operation to a 2-layer Unet. The hidden dimensions are 512, 1024, respectively. Follow\cite{DBLP:conf/nips/HoJA20}, we use SiLU as activation function and set the number of attention heads as 32.}
\revise{~\\ $\bullet$ \textit{Transformer}\cite{Peebles2022DiT}: Since there is no downsample operation for transformer based model, we here stack more layers in our model to study whether attention with large number parameters can bring better results. We instantiate our transformer-based diffusion model with 12 layers and set the hidden dimension as 1024. The number of attention heads is 32. Layer norm\cite{DBLP:journals/corr/BaKH16} is applied to this backbone, time embedding is produced by a fully-connected layer and added with AdaLN, as in\cite{Peebles2022DiT}.}

\revise{~\\ \textbf{Diffusion process}. For diffusion, we use the cosine noise schedule to produce the hyper-parameters $\{\beta_n\}_{n=1}^N$, which denote the ratio of Gaussian noise
added to the data at each diffusion step. We set diffusion step to 200 for CrossTask and COIN dataset and 50 for NIV due to their different scales. When DDIM sampling is applied, we set the sampling step number to 10 for faster inference.}

\revise{~\\ \textbf{Training}. We use the linear warm-up training scheme to optimize our model and train different steps for the three datasets, corresponding to their scales. In CrossTask$_{Base}$, we set the diffusion step as $200$ and train a Unet based model for $12,000$ steps with learning rate increasing linearly to $8 \times 10^{-4}$ in the first $4,000$ steps. Then the learning rate decays to $4 \times 10^{-4}$ at step $10,000$. In CrossTask$_{How}$, we keep diffusion step as 200 and train our Unet based model for $24,000$ steps with learning rate increasing linearly to $5 \times 10^{-4}$ in the first $4,000$ steps and decays by 0.5 at step $10,000, 16,000$ and $22,000$. When joint training is applied to CrossTask$_{How}$, only $12,000$ training steps are required since gradient descent is conducted on multiple batches with different horizons in one training step. In NIV, the diffusion step is $50$ and we train a Unet based model for $6,500$ steps due to the small size of this dataset. The learning rate increases linearly to $3 \times 10^{-4}$ in the first $4,500$ steps and decays by $0.5$ at step $6,000$. We train a Unet-attention model on the COIN dataset with task mask and two stage prediction. We set diffusion step as $200$ and train our model for $14,000$ steps.
The learning rate increases linearly to $1 \times 10^{-4}$ in the first $4,000$ steps, then we keep learning rate unchanged for the remaining training steps. The training batch size for all experiments is $256$. We set $w = 10$ for the weighted loss. All our experiments are conducted with ADAM\cite{DBLP:journals/corr/KingmaB14} on 8 NVIDIA TITAN XP GPUs.
}

\subsection{Results of task classifier}
\label{exp:task}

The first stage of our learning is to predict the task class with the given start and goal observations. We implement this with MLP models. The classification results for different planning horizons on three datasets are shown in \cref{table:classifier_cross}. We can see that our classifier can perfectly figure out the task class in the NIV dataset since only 5 tasks are involved. For larger datasets CrossTask and COIN, our model can make right predictions most of the time.

\begin{table}
\centering
\caption{Classification results on all datasets. CrossTask$_{Base}$ uses features provided by the dataset while CrossTask$_{How}$ applies features extracted by HowTo100M trained encoder.}
\resizebox*{1.0\linewidth}{!}{
\begin{tabular}{ccccc}
\hline
      & CrossTask$_{Base}$ & CrossTask$_{How}$ & COIN & NIV \\ \cline{2-5} 
$T$ = 3 &     83.87      &  92.43      &  79.42   &  100.00    \\ 
$T$ = 4 &     83.64      &  92.98      &  78.89   &  100.00    \\ 
$T$ = 5 &     83.37      &  93.39      &  -   &   -   \\ 
$T$ = 6 &     83.85      &  93.20      &  -   &   -   \\ \hline
\end{tabular}
}
\label{table:classifier_cross}
\vspace{-0.3cm}
\end{table}

\vspace{-5pt}
\subsection{Exploration Studies}
\label{exp:ablation}
To further study the effectiveness and performance of our PDPP, we in this section conduct detailed ablation studies. We start with the basic setting that we use Unet as backbone and train our model separately. Task condition is added by concatenating task feature in the conditional action sequence input. We mainly focus on the ablation study on CrossTask with  better HowTo100M features.

\subsubsection{Impact of batch size on mIoU}
\label{abla::bs}
In this section, we study the impact of batch size on mIoU, which can illustrate the importance for the standardization of computing mIoU. As we discussed in \cref{exp:protocol}, previous approaches calculate the IoU value on every mini-batch and take their mean as the final mIoU. However, the batch size value for different methods may be different, which results in an unfair comparison.

We use models trained with the basic setting and vary the evaluation batch size to compute the mIoU metric on CrossTask. We use CrossTask$_{Base}$ to denote our model with features provided by CrossTask and CrossTask$_{How}$ as model with features extracted by HowTo100M trained encoder. Planning horizon is set to $\{3, 4, 5, 6\}$. The results are shown in \cref{table3}, which validate our thought and show the huge impact of batch size on mIoU. The value of mIoU evaluated on the same model can vary widely as batch size changes, so comparing mIoU with different evaluation batch size has no meaning. To address this problem, we standardize the way to compute mIoU as setting inference batch size to 1.

\begin{table}[t]
\centering
\caption{Evaluation results of mIoU with different batch size on CrossTask.}
\resizebox*{0.9\linewidth}{!}{
\begin{tabular}{cccccc}
\hline
                      & Batch size & T = 3 & T=4 & T=5 & T=6 \\ \hline
\multirow{4}{*}{CrossTask$_{Base}$} & 1         &  56.90     &  56.99   &   56.32  & 57.51    \\
                      & 4         &  65.30     &  67.14   &  67.10   &   \textbf{70.48}  \\
                      & 8         &  68.83     &  \textbf{69.64}   &  \textbf{67.39}   &   69.31  \\
                      & 16        &   \textbf{69.79}    &   67.26  &   64.53  &  63.19   \\ \hline
\multirow{4}{*}{CrossTask$_{How}$} & 1         &   66.57    &   65.13  &   65.32  &   64.70  \\
                      & 4         &   75.21    &   77.07  &   78.56  &   78.59  \\
                      & 8         &   79.74    &   81.74  &   \textbf{81.73}  &   \textbf{80.88}  \\
                      & 16        &   \textbf{80.50}   &  \textbf{82.32}   &  81.41   &  78.64   \\ \hline
\end{tabular}
}
\label{table3}
\vspace{-0.2cm}
\end{table}

\begin{table}[]
\centering
\caption{Ablation study on the effect of joint training. HowTo100M features are applied to CrossTask.}
\resizebox*{1\linewidth}{!}{
\renewcommand\arraystretch{1.1}
\begin{tabular}{cccccccc}
\hline
\multirow{2}{*}{}    & \multicolumn{1}{l}{\multirow{2}{*}{Dataset}}   & \multicolumn{3}{c}{Train jointly} & \multicolumn{3}{c}{Train separately} \\ \cline{3-8} 
                     &                            & SR$\uparrow$         & mAcc$\uparrow$      & mIoU$\uparrow$      & SR$\uparrow$        & mAcc$\uparrow$      & mIoU$\uparrow$      \\ \hline
\multirow{3}{*}{$T$ = 3} & \multicolumn{1}{l}{CrossTask}                  & 37.00      & \revise{\textbf{65.16}}     & \revise{\textbf{67.24}}     & \textbf{37.20}      & 64.67     & 66.57     \\
                     & \multicolumn{1}{l}{NIV}                        & \textbf{30.60}      & 47.64     & \revise{\textbf{57.35}}     & 30.20      & \revise{\textbf{48.45}}     & 57.28     \\
                     & \multicolumn{1}{l}{COIN}                       & 18.21      & 43.65     & 49.11     & \textbf{21.33}     & \revise{\textbf{45.62}}     & \revise{\textbf{51.82}}     \\ \hline
\multirow{3}{*}{$T$ = 4} & \multicolumn{1}{l}{CrossTask}                  & \textbf{22.53}      & \revise{\textbf{59.09}}     & \revise{\textbf{66.57}}     & 21.48      & 57.82     & 65.13     \\
                     & \multicolumn{1}{l}{NIV}                        & \textbf{27.01}      & \revise{\textbf{46.90}}     & \revise{\textbf{60.13}}     & 26.67      & 46.89     & 59.45    \\
                     & \multicolumn{1}{l}{COIN}                       & 13.38      & 43.58     & 50.65     & \textbf{14.41}      & \revise{\textbf{44.10}}     & \revise{\textbf{51.39}}     \\ \hline
\multirow{1}{*}{$T$ = 5}    & \multicolumn{1}{l}{CrossTask} & \textbf{13.96}      & \revise{\textbf{54.64}}     & \revise{\textbf{66.43}}    & 13.45      & 54.01     & 65.32     \\
\multirow{1}{*}{$T$ = 6}    & \multicolumn{1}{l}{CrossTask} & \textbf{8.65}      & \revise{\textbf{50.46}}     & \revise{\textbf{66.18}}     & 8.41      & 49.65     & 64.70     \\ \hline
\end{tabular}}
\label{table:joint_train1}
\end{table}

\begin{table}[]
\vspace{-10pt}
\centering
\caption{Ablation study on the role of task supervision. The $w.$ $task$ $sup.$ denotes learning with task supervision and $w.o.$ $task$ $sup.$ means training with no task information.}
\resizebox*{1\linewidth}{!}{
\renewcommand\arraystretch{1.1}
\begin{tabular}{cccccccc}
\hline
\multirow{2}{*}{}    & \multicolumn{1}{l}{\multirow{2}{*}{Dataset}}   & \multicolumn{3}{c}{w. task sup.} & \multicolumn{3}{c}{w.o. task sup.} \\ \cline{3-8} 
                     &                            & SR$\uparrow$         & mAcc$\uparrow$      & mIoU$\uparrow$      & SR$\uparrow$        & mAcc$\uparrow$      & mIoU$\uparrow$      \\ \hline
\multirow{3}{*}{$T$ = 3} & \multicolumn{1}{l}{CrossTask}                  & \textbf{37.20}      & \revise{\textbf{64.67}}     & \revise{\textbf{66.57}}     & 35.69     & 63.91     & 66.04     \\
                     & \multicolumn{1}{l}{NIV}                        & \textbf{30.20}      & \revise{\textbf{48.45}}     & \revise{\textbf{57.28}}     & 28.37     & 45.96     & 54.31     \\
                     & \multicolumn{1}{l}{COIN}                       & \textbf{21.33}      & \revise{\textbf{45.62}}     & \revise{\textbf{51.82}}     & 16.48     & 36.57     & 43.48     \\ \hline
\multirow{3}{*}{$T$ = 4} & \multicolumn{1}{l}{CrossTask}                  & \textbf{21.48}      & \revise{\textbf{57.82}}     & \revise{\textbf{65.13}}     & 20.52     & 57.47     & 64.39     \\
                     & \multicolumn{1}{l}{NIV}                        & \textbf{26.67}      & \revise{\textbf{46.89}}     & \revise{\textbf{59.45}}     & 26.50     & 46.08     & 58.94     \\
                     & \multicolumn{1}{l}{COIN}                       & \textbf{14.41}      & \revise{\textbf{44.10}}     & \revise{\textbf{51.39}}     & 11.65     & 35.04     & 41.75     \\ \hline
\multirow{1}{*}{$T$ = 5}    & \multicolumn{1}{l}{CrossTask} & \textbf{13.45}      & \revise{\textbf{54.01}}     & \revise{\textbf{65.32}}    & 12.80     & 53.44     & 64.01     \\
\multirow{1}{*}{$T$ = 6}    & \multicolumn{1}{l}{CrossTask} & \textbf{8.41}      & 49.65     & \revise{\textbf{64.70}}     & 8.15     & \revise{\textbf{50.45}}     & 64.13     \\ \hline
\end{tabular}}
\label{table:task}
\end{table}

\subsubsection{Joint training vs. training separately}
\label{abla::joint_train1}
\revise{Previous approaches involved training a separate model for each prediction horizon, known as "training separately", which suffers from high training costs and complex inference. Here, we consider using data from different horizons to train one system to save training parameters, referred to as "joint training".} 
Here we study the effects of directly introducing joint training to our model. That is, we just apply joint training to the basic training setting without giving any horizon condition. \cref{table:joint_train1} presents the results, which show that joint training has a positive effect on smaller datasets CrossTask and NIV, while brings bad performance to COIN. We assume this is because much more data in large scale dataset has a more serious conflict, thus modeling the data with shared parameters can be much harder. We will introduce methods to help our model achieve better joint training results later.

\begin{figure*}[htbp]
  \centering
  \includegraphics[width=\linewidth]{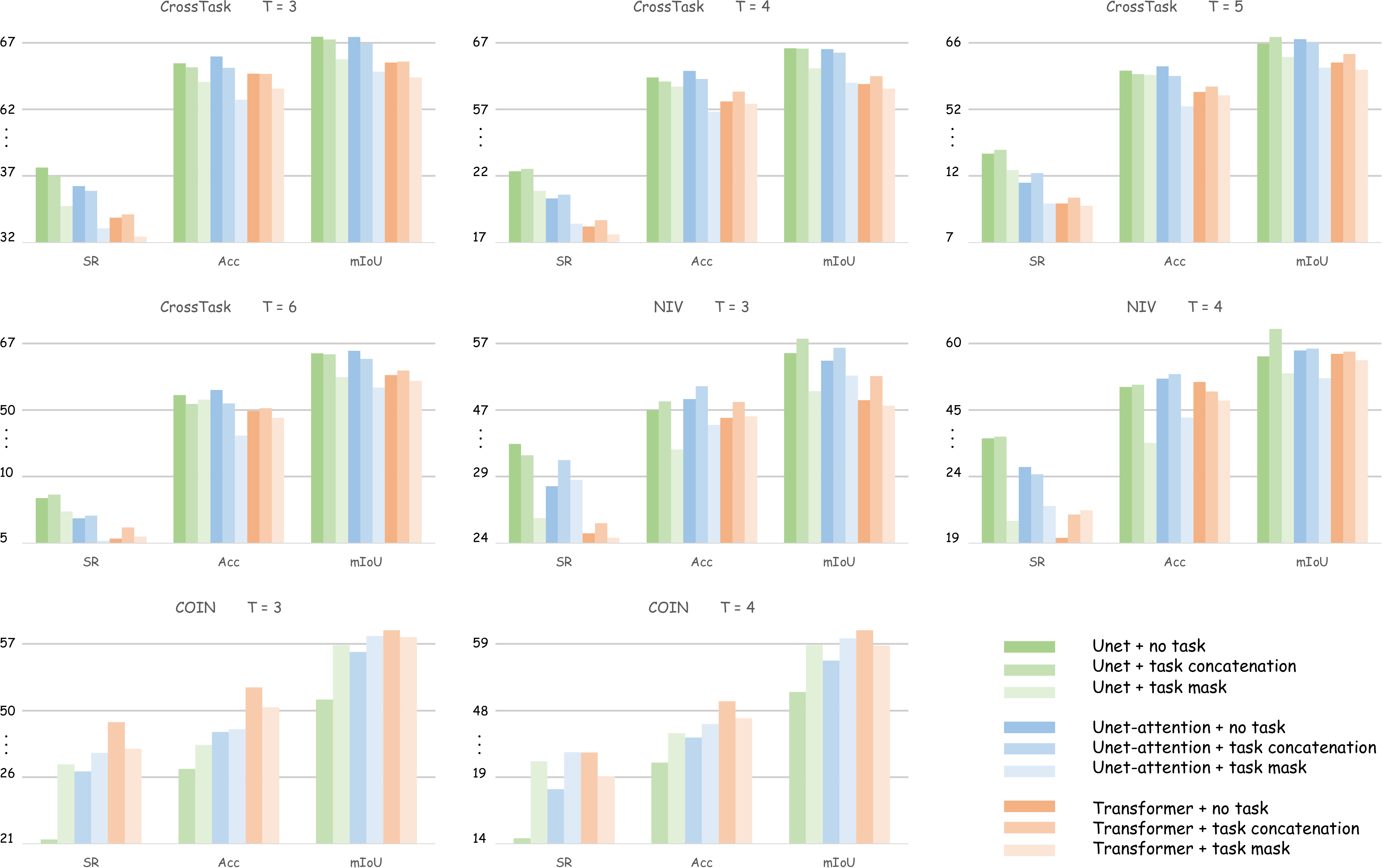} 
   \caption{\revise{Results for different model backbones with varied task adding methods on three datasets. HowTo100M features are applied to CrossTask.}}
   \label{fig:task_backbone}
   \vspace{-10pt}
\end{figure*}

\subsubsection{Study on task supervision and backbones}
\label{abla::task_backbone}
Task label plays a crucial role in our PDPP, which can provide class information about action sequence thus narrow down the searching space and help model mask better planning. In experiments we notice that the best way to add task information for different backbones can vary. So in this section, we conduct ablation studies on task supervision and backbones to get the best setting for the three datasets.

We first conduct a simple study on the role of task supervision with the basic setting by removing the task dimensions in conditional input, thus model can only plan with the start and end observations. The results are shown in \cref{table:task}. It can be seen from the results that task label is useful for all three datasets when training separately. Planning without task information brings slight performance drop on CrossTask and NIV, while all metrics drop dramatically for large scale dataset COIN. We think the reason for this is that providing task class explicitly to large scale dataset can effectively reduce the difficulty of model training, while model can learn task information of action sequence implicitly on CrossTask and NIV, thus the impact on these two datasets is much smaller.

Based on the above analysis, we try three applications of task condition on CrossTask and NIV: planning with no task, adding task information by concatenating, and multiplying task mask. For COIN dataset, task condition is crucial, so we only study the concatenation and task mask methods. Since in \cref{abla::joint_train1} we find that joint training is good for CrossTask and NIV while bad for COIN, here we just apply joint training to CrossTask and NIV and train models separately on COIN. Results for different model backbones with varied task-adding methods are shown in \cref{fig:task_backbone}.

Analyzing the study results on task supervision and backbones, we can draw the following conclusions: a) Unet based model performs best for smaller datasets and Transformer model is more suitable for large scale datasets. Task mask works bad for CrossTask and NIV but benefits COIN a lot. We believe these results are related to the dataset size. It can be expected that Transformer only performs well for COIN since applying large model to small datasets CrossTask and COIN can cause serious overfitting problem and get bad generalization results on the test set. For task mask, it is a strong planning guide which constrains the predicted results to a smaller range of actions. We assume the strong limit amplifies the distribution difference between the training and test sets. That is, task mask further breaks down the problem into learning in subspaces and reduces the difficulty of fitting the training set, thus results in overfitting problem on small datasets. b) \revise{Compared with results in \cref{table:task}, joint training makes the gap between no task and task concatenation smaller}. One explanation for this is that with joint training, more data with different horizons are provided for the model, thus model can better infer the task information from action sequences implicitly and get similar results to model trained with task supervision. c) Transformer model performs better with task concatenation than task mask on COIN dataset. One possible reason is that transformer model trained with task mask just overfits the training set of COIN.

We also notice that applying task mask to COIN can speed up the model convergence greatly. For example, when task mask is applied to Unet-attention based model on COIN, the model can converge in around 25,000 training steps. However, when concatenation is used to add task condition to model, the training step is 200,000, which is much time-consuming.

\begin{table}[t]
\centering
\caption{Ablation study on two-stage prediction process for CrossTask. We experiment with Unet models and joint training is applied.}
\resizebox*{1\linewidth}{!}{
\renewcommand\arraystretch{1.1}
\begin{tabular}{cccccccc}
\hline
\multirow{2}{*}{}    & \multicolumn{1}{l}{\multirow{2}{*}{Task sup.}}   & \multicolumn{3}{c}{w. two-stage prediction} & \multicolumn{3}{c}{w.o. two-stage prediction} \\ \cline{3-8} 
                     &                            & SR$\uparrow$         & mAcc$\uparrow$      & mIoU$\uparrow$      & SR$\uparrow$        & mAcc$\uparrow$      & mIoU$\uparrow$      \\ \hline
\multirow{2}{*}{$T$ = 3} 
                     & \multicolumn{1}{l}{no task}                        & 34.61      & 65.08     & 65.91     &  \textbf{37.64}                &              \revise{\textbf{65.47}}             &           \revise{\textbf{67.45}} \\
                     & \multicolumn{1}{l}{concatenation}                       & 33.92      & 64.61     & 65.63   &  37.00      & 65.16     & 67.24    \\ \hline
\multirow{2}{*}{$T$ = 4} 
                     & \multicolumn{1}{l}{no task}                        & 19.69      & 58.59     & 65.52     &  22.35               &            \revise{\textbf{59.39}}              &               \revise{\textbf{66.59}}   \\
                     & \multicolumn{1}{l}{concatenation}                       & 19.89      & 58.47     & 65.42     & \textbf{22.53}      & 59.09     & 66.57  \\ \hline
\multirow{2}{*}{$T$ = 5}    & \multicolumn{1}{l}{no task}                        & 12.33      & 54.46     & 65.98     &  13.68               &             \revise{\textbf{54.91}}             &                  65.92    \\
                     & \multicolumn{1}{l}{concatenation}                       & 12.08      & 53.98     & 65.89     & \textbf{13.96}      & 54.64     & \revise{\textbf{66.43}}  \\ \hline
\multirow{2}{*}{$T$ = 6}    & \multicolumn{1}{l}{no task}                        & 7.55      & 50.29     & 66.19     &  8.38            &            \revise{\textbf{51.12}}              &      \revise{\textbf{66.26}}   \\
                     & \multicolumn{1}{l}{concatenation}                       & 7.46      & 50.00     & 65.68     &  \textbf{8.65}      & 50.46     & 66.18   \\ \hline
\end{tabular}}
\label{table:a1aT1}
\end{table}

\begin{table}[t]
\centering
\caption{Ablation study on two-stage prediction process for COIN. Task mask is applied to Unet and Unet-attention models. Transformer model is trained with task concatenation. Models are trained separately.}
\resizebox*{1\linewidth}{!}{
\renewcommand\arraystretch{1.1}
\begin{tabular}{cccccccc}
\hline
\multirow{2}{*}{}    & \multicolumn{1}{l}{\multirow{2}{*}{Backbone}}   & \multicolumn{3}{c}{w. two-stage prediction} & \multicolumn{3}{c}{w.o. two-stage prediction} \\ \cline{3-8} 
                     &                            & SR$\uparrow$         & mAcc$\uparrow$      & mIoU$\uparrow$      & SR$\uparrow$        & mAcc$\uparrow$      & mIoU$\uparrow$      \\ \hline
\multirow{3}{*}{$T$ = 3} 
                     & \multicolumn{1}{l}{Unet}                        &    27.62    &   48.16   &   57.36   & 26.97     & 47.42     & 56.91     \\
                     & \multicolumn{1}{l}{Unet-attention}              & \textbf{31.05}  &  50.81  &    \revise{\textbf{59.18}}   & 27.82     & 48.61    & 57.60    \\
                     & \multicolumn{1}{l}{Transformer}                 &   29.70   &   51.53  &  58.66   & 30.14     & \revise{\textbf{51.74}}    & 58.76    \\  \hline
\multirow{3}{*}{$T$ = 4} 
                     & \multicolumn{1}{l}{Unet}                        &   21.53   &   46.67   &   59.43   & 20.20     & 46.30     & 58.92     \\
                     & \multicolumn{1}{l}{Unet-attention}              &   \textbf{22.45}  &  47.45   &   60.00  & 20.87     & 46.99     & 59.42     \\ 
                     & \multicolumn{1}{l}{Transformer}                 &   21.31  &   \revise{\textbf{49.10}}  &  \revise{\textbf{60.32}}  & 20.86      & 48.70    & 60.06      \\   \hline
\end{tabular}}
\label{table:a1aT2}
\vspace{-0.3cm}
\end{table}

\subsubsection{Study on two-stage prediction process} 
\label{abla::a1aT}
With ablation studies on task supervision and model architectures, we can select Unet backbone as the best model for CrossTask and NIV. However, for COIN dataset, though transformer based model with task concatenation performs the best, the training cost is higher than training a Unet-attention model with task mask. So in this section, we try to further split the planning process by two-stage prediction process, with an aim to achieve better performance and faster training.

The first stage in two-stage prediction strategy is to predict start and end actions with given conditions. We implement this with the same architecture of the corresponding planning model and change the learning objective to action sequence $\{a_1, a_T\}$. However, the downsample and upsample operations in Unet and Unet-attention makes action sequence of length two not available to these models. So we make some modifications by instantiating 2-layer Unet and Unet-attention models to predict action sequence $\{a_1, a_T\}$. Hidden dimensions are set to 512, 1024, respectively.

After the first prediction stage, we can consider the predicted start and end actions as additional condition and apply them to the condition projection. In experiments we find that the SR metric for predicting $\{a_1, a_T\}$ on NIV is lower than directly predicting the whole action sequence, thus using the predicted $\{a_1, a_T\}$ as condition will definitely gets worse performance. So here we just conduct experiments on CrossTask and COIN. The results for study on two-stage prediction process are shown in \cref{table:a1aT1,table:a1aT2}, in which joint training is applied only to CrossTask. As we expected in \cref{method:a1aT}, the two-stage prediction strategy provides more guidance to the sampling process and benefits performance on COIN, but causes overfitting problem on smaller datasets, thus the results on CrossTask are worse than planning the whole action sequence. Note that with the two-stage prediction process, the performance of Unet-attention model surpasses transformer based model. Besides, with the predicted $\{a_1, a_T\}$, the training of Unet-attention model can be completed in 10,000 steps, and less than 20,000 training steps are required for predicting $\{a_1, a_T\}$. Thus now we can conduct better and faster learning on COIN with Unet-attention model rather than transformer.

\begin{table}[t]
\centering
\caption{Ablation study on horizon condition for CrossTask with Unet backbone.}
\resizebox*{1\linewidth}{!}{
\renewcommand\arraystretch{1.1}
\begin{tabular}{cccccccc}
\hline
\multirow{2}{*}{}    & \multicolumn{1}{l}{\multirow{2}{*}{Task sup.}}   & \multicolumn{3}{c}{w. horizon} & \multicolumn{3}{c}{w.o. horizon} \\ \cline{3-8} 
                     &                            & SR$\uparrow$         & mAcc$\uparrow$      & mIoU$\uparrow$      & SR$\uparrow$        & mAcc$\uparrow$      & mIoU$\uparrow$      \\ \hline
\multirow{2}{*}{$T$ = 3} 
                     & \multicolumn{1}{l}{no task}                        &    \textbf{37.96}    &    \revise{\textbf{65.76}}   &   \revise{\textbf{67.75}}    &  37.64                &             65.47             &           67.45 \\
                     & \multicolumn{1}{l}{concatenation}                       &   37.46     &   65.34    &  67.35   &  37.00      & 65.16     & 67.24    \\ \hline
\multirow{2}{*}{$T$ = 4} 
                     & \multicolumn{1}{l}{no task}                        &   22.56   &   \revise{\textbf{59.43}}   &  \revise{\textbf{66.72}}   &  22.35               &            59.39              &               66.59   \\
                     & \multicolumn{1}{l}{concatenation}                       &  \textbf{22.83}   &   59.22    &  66.39    & 22.53      & 59.09     & 66.57  \\ \hline
\multirow{2}{*}{$T$ = 5}    & \multicolumn{1}{l}{no task}                        &  \textbf{14.30}    &   \revise{\textbf{55.17}}   &   \revise{\textbf{66.68}}   &  13.68               &             54.91             &                  65.92    \\
                     & \multicolumn{1}{l}{concatenation}                       &  13.79   &   54.58   &   66.19   & 13.96      & 54.64     & 66.43  \\ \hline
\multirow{2}{*}{$T$ = 6}    & \multicolumn{1}{l}{no task}                        &  \textbf{8.93}   &  \revise{\textbf{51.45}}   &  \revise{\textbf{66.56}} &  8.38            &            51.12              &      66.26   \\
                     & \multicolumn{1}{l}{concatenation}                       &   8.65   &   50.47   &  66.11   &  8.65      & 50.46     & 66.18   \\ \hline
\end{tabular}}
\label{table:moe1}
\end{table}

\begin{table}[t]
\centering
\caption{Ablation study on horizon condition \revise{for NIV} with Unet backbone.}
\resizebox*{1\linewidth}{!}{
\renewcommand\arraystretch{1.1}
\begin{tabular}{cccccccc}
\hline
\multirow{2}{*}{}    & \multicolumn{1}{l}{\multirow{2}{*}{Task sup.}}   & \multicolumn{3}{c}{w. horizon} & \multicolumn{3}{c}{w.o. horizon} \\ \cline{3-8} 
                     &                            & SR$\uparrow$         & mAcc$\uparrow$      & mIoU$\uparrow$      & SR$\uparrow$        & mAcc$\uparrow$      & mIoU$\uparrow$      \\ \hline
\multirow{2}{*}{$T$ = 3} 
                     & \multicolumn{1}{l}{no task}                        &   29.18  &   45.21  &  55.08   &  \textbf{31.46}                &  47.00  &  56.27 \\
                     & \multicolumn{1}{l}{concatenation}                       &  30.74   &  \revise{\textbf{48.10}}   & \revise{\textbf{57.96}}  &  30.60   & 47.64  & 57.35   \\ \hline
\multirow{2}{*}{$T$ = 4} 
                     & \multicolumn{1}{l}{no task}                        &   26.15   &   46.74   &   59.58    &  26.87              &  46.72  &   58.02   \\
                     & \multicolumn{1}{l}{concatenation}                       &   \textbf{27.78}   &   46.85  &  59.62    & 27.01     & \revise{\textbf{46.90}}   & \revise{\textbf{60.13}} \\ \hline
\end{tabular}}
\label{table:moe2}
\vspace{-0.3cm}
\end{table}

\begin{table}[]
\vspace{-10pt}
\centering
\caption{Ablation study on horizon for COIN with Unet-attention backbone and task mask. Two stage prediction process is applied. MoEs\_atten and MoEs\_conv denote applying MoEs to all attention or convolution layers.}
\resizebox*{1\linewidth}{!}{
\renewcommand\arraystretch{1.1}
\begin{tabular}{cccccccc}
\hline
\multirow{2}{*}{Horizon sup.}    & \multicolumn{1}{l}{\multirow{2}{*}{Routing}}   & \multicolumn{3}{c}{$T$ = 3} & \multicolumn{3}{c}{$T$ = 4} \\ \cline{3-8} 
                     &                            & SR$\uparrow$         & mAcc$\uparrow$      & mIoU$\uparrow$      & SR$\uparrow$        & mAcc$\uparrow$      & mIoU$\uparrow$      \\ \hline
\multirow{1}{*}{\color{gray}{Train separately}} 
                     & \multicolumn{1}{c}{\color{gray}{-}}                        &    \color{gray}{31.05}    &   \color{gray}{50.81}   &   \color{gray}{59.18}   & \color{gray}{22.45}     & \color{gray}{47.45}     & \color{gray}{60.00}     \\ \hline
\multirow{1}{*}{No horizon} 
                     & \multicolumn{1}{c}{-}                        &    28.97    &   50.58   &   58.35   & 21.95     & 48.06     & 60.16     \\ \hline
\multirow{1}{*}{Concatenation}
                    & \multicolumn{1}{c}{-}              & \textbf{30.12}  &  \revise{\textbf{50.95}}  &    \revise{\textbf{59.00}}   & 22.24     & 48.15    & 60.22    \\ \hline
\multirow{2}{*}{MoEs\_atten} 
                     & \multicolumn{1}{l}{Direct}                        &   29.56   &   50.70   &   58.46   & 22.25     & 48.17     & \revise{\textbf{60.31}}     \\
                     & \multicolumn{1}{l}{Learned}              &   29.64  &  50.82   &   58.53  & 21.98     & 48.08     & 60.17     \\ \hline
\multirow{2}{*}{MoEs\_conv} 
                     & \multicolumn{1}{l}{Direct}                        &   29.44   &   50.61   &   58.37   & \textbf{22.46}     & \revise{\textbf{48.25}}     & 60.04     \\
                     & \multicolumn{1}{l}{Learned}              &   29.47  &  50.70   &   58.35  & 22.23     & 48.22     & 60.26     \\ \hline
\end{tabular}}
\label{table:moe3}
\end{table}

\subsubsection{Study on methods for better joint training}
\label{abla::joint_train2}
We in this section study two methods to utilize horizon information for better joint training: horizon concatenation and MoEs. We try horizon concatenation on all datasets. For MoEs, as explained in \cref{method:diffusion}, we only apply it to the convolution part or attention part of Unet-attention model with direct routing and learned routing strategies on COIN since Unet-attention model only performs well on large scale dataset. The results are presented in \cref{table:moe1,table:moe2,table:moe3}.

For CrossTask, we can see that adding horizon condition is overall positive to our model, especially for models trained with no task. For NIV, applying horizon information along with task concatenation can benefit models, while harms the performance of models with no task. As for COIN dataset, joint training still has a negative impact on our model. Compared with directly train PDPP on COIN jointly("No horizon" in \cref{table:moe3}), utilizing horizon concatenation and applying MoEs can both alleviate the influence caused by joint training. On the whole, horizon concatenation performs the best and requires the minimum additional training parameters.

With all above ablation studies, we can now get the best settings for our joint trained PDPP on different datasets. For CrossTask, Unet based model trained with horizon condition concatenation is applied. Task supervision is not required. For NIV, we just concatenate task and horizon conditions to the action sequence to train our Unet diffusion model. For COIN, we use the Unet-attention backbone to implement our model. Horizon information is added by concatenation and task condition is added by task mask. Two stage prediction is applied for COIN.
\vspace{-8pt}

\re{\subsubsection{Study on conditions applied to PDPP}}
\label{abla::condition_PDPP}
\noindent \re{In this section we conduct a more comprehensive study on conditions applied to PDPP. We chose to experiment on the largest and most comprehensive dataset COIN to achieve a thorough evaluation.}

\re{We first ablate on sampling conditions to figure out how different conditions affect PDPP. Apart from predicted task class ($task$), predicted start-end actions ($a_1a_T$) and given observations ($ob.$) mentioned in the article, we also follow Stable Diffusion\cite{stable_diff} to encode the name of predicted task with CLIP\cite{CLIP} text encoder as language condition ($lang.$). \cref{table:ablation_cond} below shows the ablation results. It can be seen that the input observations and start-end actions predicted by given observations are important for the successful prediction of PDPP, which is natural because without observations (see the first row in \cref{table:ablation_cond}), we can only randomly predict action sequences from the corresponding category task. Although this sequence may maintain the logical relationship between actions, it would not correspond to the given observations. Moreover, introducing language embedding does not improve the prediction effect. Useful conditions are the given observations, predicted actions and tasks. We think this is because in a non-open-vocabulary prediction setting, the main role of conditions is to guide model to distinguish between different tasks and actions, thus using mask or one-hot embedding as condition is more direct and easier to learn than introducing language embeddings.}

\begin{table}[t]
\centering
\caption{\re{Ablation study on conditions applied to PDPP \revise{with the COIN dataset}.}}
\resizebox*{1.0\linewidth}{!}{
\re{
\begin{tabular}{ccccccclll}
\hline
\multirow{2}{*}{$ob.$} & \multirow{2}{*}{$a_1a_T$} & \multirow{2}{*}{$task$} & \multirow{2}{*}{$lang.$} & \multicolumn{3}{c}{T=3} & \multicolumn{3}{c}{T=4}                                                      \\ \cline{5-10} 
& & & & SR$\uparrow$    & mAcc$\uparrow$   & mIoU$\uparrow$   & \multicolumn{1}{c}{SR$\uparrow$} & \multicolumn{1}{c}{mAcc$\uparrow$} & \multicolumn{1}{c}{mIoU$\uparrow$} \\ \hline
\xmark & \xmark & \cmark & \xmark & 0.11 & 0.45 & 0.70 & 0.06 & 0.52 & 0.82 \\
\xmark & \cmark & \cmark & \xmark & 27.79 & 49.91 & 56.83 & 16.74 & 44.11 & 54.32 \\
\cmark & \xmark & \cmark & \xmark & 28.85 & 50.06 & 58.07 & 21.38 & 48.05 & 59.86 \\
\cmark & \cmark & \cmark & \xmark & \textbf{30.12} & 50.95 & \revise{\textbf{59.00}} & \textbf{22.24} & 48.15 & \revise{\textbf{60.22}} \\ \hline
\cmark & \cmark & \xmark & \xmark & 28.41 & \revise{\textbf{51.35}} & 58.17 & 19.87 & \revise{\textbf{48.68}} & 59.52 \\
\cmark & \cmark & \xmark & \cmark & 28.10 & 50.84 & 58.01 & 19.68 & 47.97 & 59.29 \\
\cmark & \cmark & \cmark & \cmark & 29.77 & 50.90 & 58.90 & 21.58 & 47.78 & 60.21 \\ \hline                        
\end{tabular}
}
}
\label{table:ablation_cond}
\vspace{-5pt}
\end{table}

\re{
Based on above findings, we retained the effective conditions and studied the effect of applying the Classifier-Free Guidance (CFG) strategy to PDPP. Specifically, we follow\cite{DBLP:journals/corr/abs-2207-12598} to train PDPP by removing a certain proportion (20\%) of conditions, allowing it to perform both conditional and unconditional predictions simultaneously. After trained, we sample from the learned distribution with the following formula:
\begin{equation}
\epsilon_t = \epsilon_\theta(x_t, c) + \lambda(\epsilon_\theta(x_t, c) - \epsilon_\theta(x_t))
\end{equation}
where $c$ denotes conditions and $\lambda$ controls the degree to which the sampling is influenced by the condition. We then conducted ablation on the value of $\lambda$. Since predicted start-end actions from the first stage are part of the prediction result, we combine them with randomly selected intermediate actions as a baseline for comparison. The results provided in \cref{table:ablation_cfg} show that: (1) Even without any conditions ($\lambda = -1$), PDPP can learn some of the logical relationships between actions and thus obtain better results than the random baseline; (2) A smaller $\lambda(0-1)$ slightly improves the planning result, while larger $\lambda(> 1)$ can cause prediction errors. (3) Removing a certain proportion of conditions during training is a data augmentation method that can improve model performance on COIN. Overall, the CFG strategy provides little improvement for the effectiveness of PDPP. We believe this is due to the semantic differences between procedure planning and image generation. Unconditional prediction, lacking observations, cannot achieve accurate plans and tends to predict action sequences that have a higher probability of occurring. Thus using the difference between conditional and unconditional predictions as the main factor ($\lambda>1$) would result in poorer results.
}

\begin{table}[t]
\centering
\caption{\re{Ablation study of the cfg\_scale value $\lambda$ on the largest COIN dataset. \revise{Two-stage prediction process is applied.}}}
\vspace{-5pt}
\resizebox*{1.0\linewidth}{!}{
\re{
\begin{tabular}{ccccccc}
\hline
                   & \multicolumn{3}{c}{T=3} & \multicolumn{3}{c}{T=4} \\ \cline{2-7} 
                   & SR$\uparrow$    & mAcc$\uparrow$   & mIoU$\uparrow$   & SR$\uparrow$    & mAcc$\uparrow$   & mIoU$\uparrow$   \\ \hline
Random Baseline    &    9.59   &    42.63    &    47.22    &   2.27    &    36.31    &    47.75    \\
$\lambda$=-1.0(uncondition) &   18.54    &    47.55    &    53.47    &  13.08     &   45.74     &    56.50    \\
$\lambda$=0.0(condition)    &   30.33    &    \revise{\textbf{51.16}}    &    58.99    &    22.60   &   48.59     &     60.32   \\
$\lambda$=0.1               &   \textbf{30.34}    &    \revise{\textbf{51.16}}    &    \revise{\textbf{59.00}}    &    22.60   &   48.58     &     60.34   \\
$\lambda$=0.3               &   30.33    &    51.15    &    58.99    &    \textbf{22.75}   &   \revise{\textbf{48.62}}     &     60.38   \\
$\lambda$=0.5               &   30.31    &    51.12    &    58.97    &    22.72   &   48.59     &     \revise{\textbf{60.39}}   \\
$\lambda$=1.0               &   30.21    &    51.08    &    58.93    &    22.52   &   48.45     &     60.31   \\
$\lambda$=2.0               &   30.01    &   51.00     &   58.83     &    21.97   &   48.23     &     60.09   \\
$\lambda$=3.0               &   29.74    &   50.90     &   58.70     &    21.09   &   47.87     &     59.76   \\
$\lambda$=4.0               &   29.40    &   50.76     &   58.53     &    19.80   &   47.36     &     59.21   \\
$\lambda$=5.0               &   28.90    &   50.55     &   58.27     &    18.44   &   46.81     &     58.56   \\ \hline
\end{tabular}
}
}
\label{table:ablation_cfg}
\vspace{-5pt}
\end{table}

\vspace{3pt}
\subsection{Comparison with other approaches}
\label{exp:compare}

\begin{table*}[t]
\centering
\caption{Evaluation results on CrossTask for procudure planning with prediction horizon $T \in \{3, 4\}$. The $Feature$ column denotes the type of video feature used for learning. The $Supervision$ column denotes the type of supervision applied in training, where V denotes intermediate visual states, L denotes language feature and C means task class. \textbf{Note that we compute mIoU by calculating average of every IoU of a single antion sequence rather than a mini-batch}.}
\resizebox*{0.85\linewidth}{!}{
\begin{tabular}{ccccccccc}
\hline
       &          &             & \multicolumn{3}{c}{$T$ = 3} & \multicolumn{3}{c}{$T$ = 4} \\ \cline{4-9} 
\multicolumn{1}{l}{Models}    & \multicolumn{1}{l}{Feature}       & Supervision & SR$\uparrow$    & mAcc$\uparrow$   & mIoU$\uparrow$   & SR$\uparrow$    & mAcc$\uparrow$   & mIoU$\uparrow$   \\ \hline
\multicolumn{1}{l}{Random}    & \multicolumn{1}{l}{Base}       &     -        &   $<$0.01    &    0.94    &   \color{gray}{1.66}     &   $<$0.01    &    0.83    &   \color{gray}{1.66}     \\
\multicolumn{1}{l}{Retrieval-Based}  & \multicolumn{1}{l}{Base}  &     -        &   8.05    &   23.30     &   \color{gray}{32.06}     &   3.95    &    22.22    &    \color{gray}{36.97}    \\
\multicolumn{1}{l}{WLTDO\cite{DBLP:conf/cvpr/EhsaniBRMF18}}      & \multicolumn{1}{l}{Base}       &      -       &   1.87    &    21.64    &   \color{gray}{31.70}     &   0.77    &    17.92    &    \color{gray}{26.43}    \\
\multicolumn{1}{l}{UAAA\cite{DBLP:conf/iccvw/FarhaG19}}      & \multicolumn{1}{l}{Base}        &      -       &   2.15    &   20.21     &   \color{gray}{30.87}     &   0.98    &   19.86     &  \color{gray}{27.09}      \\
\multicolumn{1}{l}{UPN\cite{DBLP:conf/icml/SrinivasJALF18}}     & \multicolumn{1}{l}{Base}          &      V       &   2.89    &    24.39    &   \color{gray}{31.56}     &   1.19    &   21.59     &   \color{gray}{27.85}     \\
\multicolumn{1}{l}{DDN\cite{DBLP:conf/eccv/ChangHXAFN20}}      & \multicolumn{1}{l}{Base}         &      V       &   12.18    &    31.29    &    \color{gray}{47.48}    &   5.97    &    27.10    &    \color{gray}{48.46}    \\
\multicolumn{1}{l}{Ext-GAILw/o Aug.\cite{DBLP:conf/iccv/BiLX21}}  & \multicolumn{1}{l}{Base}  &      V       &   18.01    &   43.86     &   \color{gray}{57.16}     &    -   &    -    &    -    \\
\multicolumn{1}{l}{Ext-GAIL\cite{DBLP:conf/iccv/BiLX21}}    & \multicolumn{1}{l}{Base}      &      V       &   21.27    &   49.46     &   \color{gray}{61.70}     &   16.41    &    43.05    &   \color{gray}{60.93}     \\
\multicolumn{1}{l}{Ours$_{Base}$}     & \multicolumn{1}{l}{Base}         &      C       &    26.43   &    55.04    &     \color{gray}{57.93}   &   16.20    &   50.85     &    \color{gray}{58.47}    \\ \hline
\multicolumn{1}{l}{P$^3$IV\cite{DBLP:conf/cvpr/0004HDDWJ22}}     & \multicolumn{1}{l}{HowTo100M}          &      L       &   23.34    &   49.96     &   \color{gray}{73.89}     &   13.40    &   44.16     &   \color{gray}{70.01}     \\

\multicolumn{1}{l}{Ours$_{How}$}     & \multicolumn{1}{l}{HowTo100M}         &      -       &    \textbf{37.96}   &    \textbf{65.76}    &     \color{gray}{67.75}   &   \textbf{22.56}    &   \textbf{59.43}     &    \color{gray}{66.72}    \\
\hline
\end{tabular}
}

\label{table:cross_34}
\end{table*}

In this section, We follow previous work\cite{DBLP:conf/cvpr/0004HDDWJ22} and compare our approach with other alternative methods for procedure planning on the three datasets, across multiple prediction horizons.

\vspace{-5pt}
~\\ \noindent \textbf{Baselines.} 

\noindent - $Random$. This policy randomly selects actions from the available action space in dataset to produce the plans.

\noindent - $Retrieval$-$Based$. Given the observations $\{o_s, o_g\}$, the retrieval-based method retrieves the closest neighbor by calculating the minimum visual feature distance in the train dataset. Then the action sequence associated with the retrieved result will be used as the action plan.

\noindent - $WLTDO$\cite{DBLP:conf/cvpr/EhsaniBRMF18}. This approach applies a recurrent neural network(RNN) to predict action steps with the given observation pairs.

\noindent - $UAAA$\cite{DBLP:conf/iccvw/FarhaG19}. UAAA is a two-stage approach which uses RNN-HMM model to predict action steps autoregressively.

\noindent - $UPN$\cite{DBLP:conf/icml/SrinivasJALF18}. UPN is a
physical-world path planning algorithm and learns a plannable representation to make predictions. To produce the discrete action steps, we follow\cite{DBLP:conf/eccv/ChangHXAFN20} to add a softmax layer to the output of this model.

\noindent - $DDN$\cite{DBLP:conf/eccv/ChangHXAFN20}. DDN model is a two-branch autoregressive model, which learns an abstract representation of action steps and tries to predict the state-action transition in the feature space.

\noindent - $PlaTe$\cite{DBLP:journals/ral/SunHLLZG22}. PlaTe model follows DDN and uses transformer modules in two-branch to predict instead. Note that the evaluation protocol of PlaTe is different with other models, so we move the comparison with PlaTe to supplementary material, which we will discuss later.

\noindent - $Ext$-$GAIL$\cite{DBLP:conf/iccv/BiLX21}. This model solves the procedure planning problem by reinforcement learning techniques. Similar to our work, Ext-GAIL decomposes the procedure planning problem into two sub-problems. However, the purpose of the first sub-problem in Ext-GAIL is to provide long-horizon information for the second stage while our purpose is to get condition for sampling.

\noindent - $P^3IV$\cite{DBLP:conf/cvpr/0004HDDWJ22}. P$^3$IV is a single-branch transformer-based model which augments itself with a learnable memory bank and an extra generative adversarial framework. Like our model, P$^3$IV predicts all action steps at once during inference process.

~\\ \noindent \textbf{CrossTask (short horizon).} 
We first evaluate on CrossTask with two prediction horizons typically used in previous work. We use Ours$_{Base}$ to denote our model with features provided by CrossTask and Ours$_{How}$ as model with features extracted by HowTo100M trained encoder. For Ours$_{Base}$, here we apply Unet model trained jointly with task and horizon concatenation to compare with previous methods. Note that we compute mIoU by calculating the mean of every IoU for a single antion sequence rather than a mini-batch as explained in \cref{exp:protocol}, though the latter can achieve a higher mIoU value. Results in \cref{table:cross_34} show that Ours$_{Base}$ beats all methods for most metrics except for the success rate (SR) when $T = 4$, where our model is the second best, and Ours$_{How}$ just significantly outperforms all previous methods. Specifically, for using HowTo100M-extracted video features, we outperform\cite{DBLP:conf/cvpr/0004HDDWJ22} by more than 14\% and 9\% for SR when $T=3,4$, respectively. As for features provided by CrossTask, Ours$_{Base}$ outperforms the previous best method\cite{DBLP:conf/iccv/BiLX21} by more than 5\% both for SR and mAcc when $T=3$.

\begin{table}[t]
\centering
\caption{$SR$ evaluation results on CrossTask with longer planning horizons.}
\resizebox*{0.8\linewidth}{!}{
\renewcommand\arraystretch{1.1}
\begin{tabular}{ccccc}
\hline
                & $T$ = 3   & $T$ = 4   & $T$ = 5  & $T$ = 6  \\ \cline{2-5} 
\multicolumn{1}{l}{Models}          & SR$\uparrow$    & SR$\uparrow$    & SR$\uparrow$   & SR$\uparrow$   \\ \hline
\multicolumn{1}{l}{Retrieval-Based} & 8.05  & 3.95  & 2.40 & 1.10 \\
\multicolumn{1}{l}{DDN\cite{DBLP:conf/eccv/ChangHXAFN20}}             & 12.18 & 5.97  & 3.10 & 1.20 \\
\multicolumn{1}{l}{P$^3$IV\cite{DBLP:conf/cvpr/0004HDDWJ22}}            & 23.34 & 13.40 & 7.21 & 4.40 \\
\multicolumn{1}{l}{Ours$_{Base}$}            & 26.43 & 16.20 & 9.20 & 6.28 \\
\multicolumn{1}{l}{Ours$_{How}$}            & \textbf{37.96} & \textbf{22.56} & \textbf{14.30} & \textbf{8.93} \\
\hline
\end{tabular}
}
\label{table:cross_56}
\vspace{-5pt}
\end{table}

~\\ \noindent \textbf{CrossTask (long horizon).} We further study the ability of predicting with longer horizons for our model. Following\cite{DBLP:conf/cvpr/0004HDDWJ22}, we here evaluate the SR value with planning horizon $T = \{3, 4, 5, 6\}$. We present the result of our model along with other approaches that reported results for longer horizons in \cref{table:cross_56}. This result shows our model can get a stable and great improvement with all planning horizons compared with the previous best model.

\begin{table}[b]
\vspace{-10pt}
\centering
\caption{Evaluation results on NIV and COIN with prediction horizon $T \in \{3,4\}$. Sup. denotes the type of supervision in training. Note that we compute IoU on every action sequence and take the mean as mIoU.} 
\resizebox*{1.0\linewidth}{!}{
\renewcommand\arraystretch{1}
\begin{tabular}{cccccccc}
\hline
                     &                      & \multicolumn{3}{c}{NIV} & \multicolumn{3}{c}{COIN} \\ \cline{3-8} 
Horizon              & \multicolumn{1}{l}{Models}         
& SR$\uparrow$     & mAcc$\uparrow$   & mIoU$\uparrow$  & SR$\uparrow$     & mAcc$\uparrow$   & mIoU$\uparrow$   \\ \hline
\multirow{6}{*}{$T$ = 3} & \multicolumn{1}{l}{Random}          
& 2.21   & 4.07   & \color{gray}{6.09}  &   $<$0.01     &    $<$0.01    & \color{gray}{2.47}   \\
                     & \multicolumn{1}{l}{Retrieval} 
                     & -      & -      & -     & 4.38   & 17.40  & \color{gray}{32.06}  \\
                     & \multicolumn{1}{l}{DDN\cite{DBLP:conf/eccv/ChangHXAFN20}}            
                     & 18.41  & 32.54  & \color{gray}{56.56} & 13.9   & 20.19  & \color{gray}{64.78}  \\
                     & \multicolumn{1}{l}{Ext-GAIL\cite{DBLP:conf/iccv/BiLX21}}       
                     & 22.11  & 42.20  & \color{gray}{65.93} & -      & -      & -      \\
                     & \multicolumn{1}{l}{P$^3$IV\cite{DBLP:conf/cvpr/0004HDDWJ22}}           
                     & 24.68  & \textbf{49.01}  & \color{gray}{74.29} & 15.4   & 21.67  & \color{gray}{76.31}  \\
                     & \multicolumn{1}{l}{Ours}            
                     & \textbf{30.74}  & 48.10  & \color{gray}{57.96} & \textbf{30.12}  & \textbf{50.95}  & \color{gray}{59.00}  \\ \hline
\multirow{6}{*}{$T$ = 4} & \multicolumn{1}{l}{Random}          
& 1.12   & 2.73   & \color{gray}{5.84}  &   $<$0.01     &    $<$0.01    & \color{gray}{2.32}   \\
                     & \multicolumn{1}{l}{Retrieval} 
                     & -      & -      & -     & 2.71   & 14.29  & \color{gray}{36.97}  \\
                     & \multicolumn{1}{l}{DDN\cite{DBLP:conf/eccv/ChangHXAFN20}}            
                     & 15.97  & 27.09  & \color{gray}{53.84} & 11.13  & 17.71  & \color{gray}{68.06}  \\
                     & \multicolumn{1}{l}{Ext-GAIL\cite{DBLP:conf/iccv/BiLX21}}       
                     & 19.91  & 36.31  & \color{gray}{53.84} & -      & -      & -      \\
                     & \multicolumn{1}{l}{P$^3$IV\cite{DBLP:conf/cvpr/0004HDDWJ22}}           
                     & 20.14  & 38.36  & \color{gray}{67.29} & 11.32  & 18.85  & \color{gray}{70.53}  \\
                     & \multicolumn{1}{l}{Ours}            
                     & \textbf{27.78}  & \textbf{46.85}  & \color{gray}{59.62} & \textbf{22.24}  & \textbf{48.15}  & \color{gray}{60.22}  \\ \hline
\end{tabular}
}
\label{table:coin_niv}

\end{table}

~\\ \noindent \textbf{NIV and COIN.} \cref{table:coin_niv} shows our evaluation results on the other two datasets NIV and COIN, from which we can see that our approach remains to be the best performer for both datasets. Specifically, in the NIV dataset where mAcc is relatively high, our model raises the SR value by more than 6\% both for the two horizons and outperforms the previous best by more than 8\% on mAcc metric when $T=4$. As for the large COIN dataset where mAcc is low, our model significantly improves mAcc by around 30\% and almost double the SR value when $T=3, 4$. 

All the results suggest that our model performs well across datasets with different scales.

\subsection{\re{Analysis of Failure Cases}}
\label{exp:error}

\begin{figure}[t]
  \centering
  \includegraphics[width=1.0\linewidth]{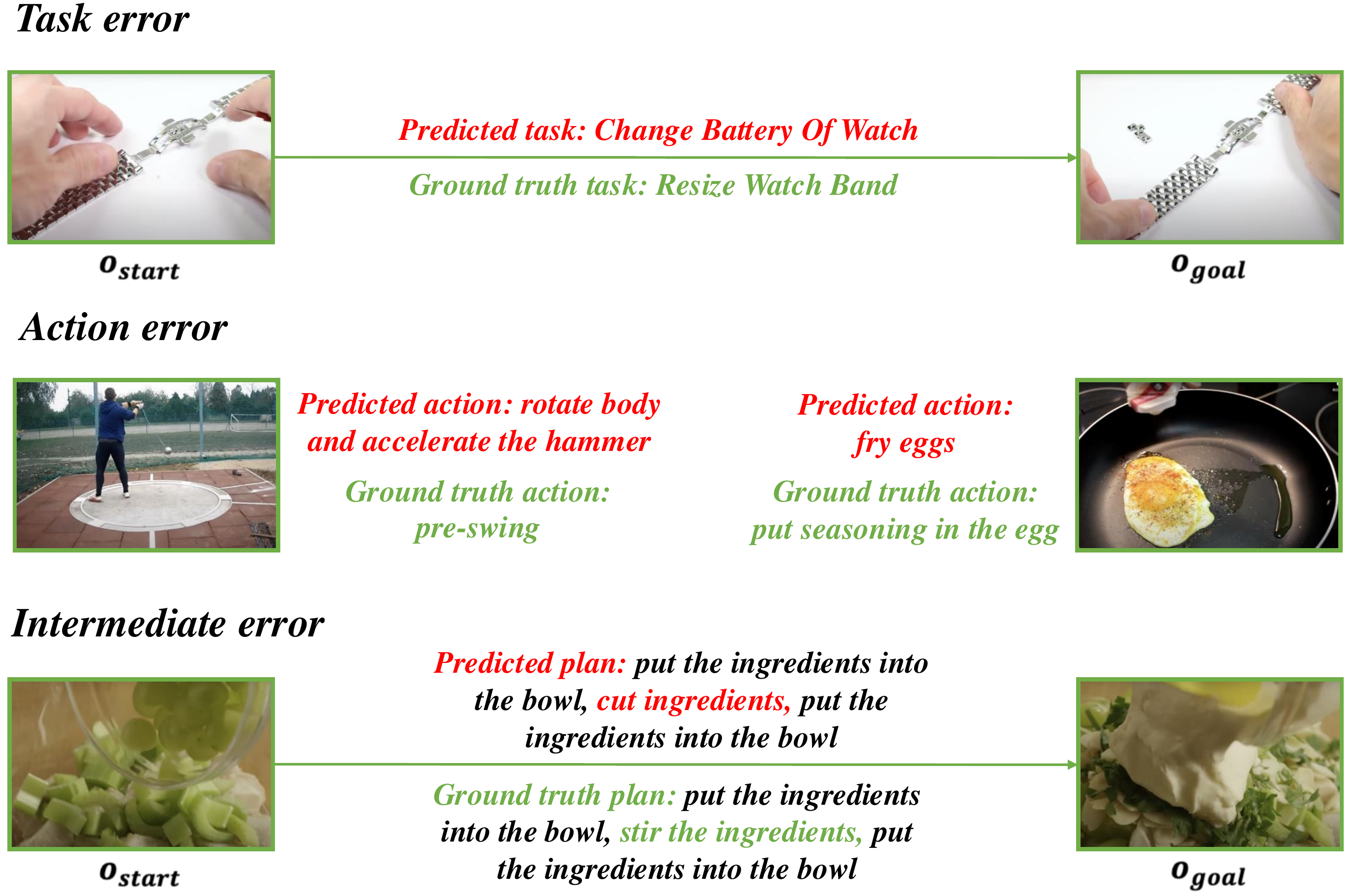} 
   \caption{\re{Visualization of task, action and intermediate errors.}}
   \label{fig:error}
\end{figure}

\re{We further analyze failure cases where PDPP makes wrong predictions on the largest and most comprehensive COIN dataset. Specifically, we calculate the proportions of the following three types of errors: task prediction error (task error), start-end actions prediction error (action error), and intermediate action prediction error (intermediate error). The results are presented in \cref{table:error_statistic}. Visualization results are shown in \cref{fig:error}. It can be seen that the primary cause of failure cases is start-end actions prediction error, indicating that inferring semantic action information from given observations is the key bottleneck to be improved. Model needs to incorporate finer details observed, such as the lower hammer and seasoning bottle in \cref{fig:error} ``Action error'', to make more accurate predictions. For task predicting error, an imbalance in the distribution of task categories within the training set makes certain tasks difficult to recognize accurately (e.g. very few training samples of task ``Resize Watch Band''). Besides, similarities between certain tasks give rise to errors in prediction due to their overlapping nature (e.g. ``Make Cookie'' and ``Make Chocolate'' share similar steps, ``Change Battery Of Watch'' and ``Resize Watch Band'' share similar appearances). Intermediate errors are caused by two reasons: the uncertainty of procedure planning and the gap between training and inference. In the example of \cref{fig:error}, all the ingredients in the test video have already been cut, but the model has encountered many examples during training where ingredients are chopped and then added into the bowl, so it predicts the action of ``cut ingredients''.
}

\begin{table}[t]
\centering
\caption{\re{Statistical results of different error types on the largest COIN dataset.}}
\resizebox*{1.0\linewidth}{!}{
\re{
\begin{tabular}{cccc}
\hline
      & task error & action error & intermediate error \\ \cline{1-4} 
$T$ = 3 &     29.6\%      &  56.3\%      &  14.1\%   \\ 
$T$ = 4 &     26.9\%      &  47.7\%      &  25.4\%   \\ \hline
\end{tabular}
}
}
\label{table:error_statistic}
\end{table}

\begin{table}[t]
\centering
\caption{\re{Computational efficiency study of PDPP. The $w. $ $j.t.$
denotes learning with joint training and $w.o.$ $ j.t.$ means
training separately.}}
\resizebox*{1.0\linewidth}{!}{
\re{
\begin{tabular}{ccccccc}
\hline
\multirow{2}{*}{Dataset} & \multirow{2}{*}{FLOPs(M)} & \multirow{2}{*}{time(s)} & \multicolumn{2}{c}{Training Params(M)} & \multicolumn{2}{c}{Training steps} \\ \cline{4-7} 
&      &       & w. j.t. & w.o. j.t. & w. j.t. & w.o. j.t.\\ \hline
CrossTask &   52.98   &   0.13    &   41.78   &  41.77 * $T$   &   24,000   & 24,000 * $T$ \\
NIV &   52.77   &   0.12    &   41.71   &   41.71 * $T$   &   6,500   & 6,500 * $T$ \\
COIN &  341.54    &   0.26    &   157.44   &  157.42 * $T$   &   14,000   & 14000 * $T$ \\ \hline               
\end{tabular}
}
}
\label{table:efficiency}
\end{table}

\begin{table}[t]
\centering
\caption{Evaluation results of probabilistic modeling metrics on COIN and NIV. Joint training with horizon concatenation is applied. Unet-attention based model with task mask and Unet based model with task concatenation are used for COIN and NIV, respectively.}
\resizebox*{0.95\linewidth}{!}{
\renewcommand\arraystretch{1}
\begin{tabular}{cccccc}
\hline
& & \multicolumn{2}{c}{COIN} & \multicolumn{2}{c}{NIV} \\ \cline{3-6}
Metric & Model  & $T$ = 3         & $T$ = 4           & $T$ = 3         & $T$ = 4 \\ \hline
\multirow{3}{*}{NLL$\downarrow$}  & Deterministic &  5.29  &  5.69  &  5.53   &  5.57  \\
                     & Noise         &  5.15  &  5.46  &   5.11   &   5.13   \\
                     & Ours          &  \textbf{5.03}  &  \textbf{5.20}  &   \textbf{4.77}   &   \textbf{4.58}  \\ \hline
\multirow{3}{*}{KL-Div$\downarrow$} & Deterministic &  4.34  &  4.29  &  5.45  &   5.44   \\
                     & Noise   &  4.21  &  4.06  &   5.03   &  5.01  \\
                     & Ours   &  \textbf{4.08}  &  \textbf{3.79}  &   \textbf{4.69}   &  \textbf{4.46}  \\ \hline
\multirow{3}{*}{SR$\uparrow$}       & Deterministic &  \textbf{30.70}  &  \textbf{23.39}  &  27.94    &  24.14   \\
                          & Noise         &  30.66  &  23.02  &   25.36   &   23.83    \\
                          & Ours          &   30.12 &  22.24  &   \textbf{30.74}   &  \textbf{27.78}  \\ \hline
\multirow{3}{*}{ModePrec$\uparrow$} & Deterministic &  \textbf{37.89}  &  \textbf{35.64}  &   30.00   &  25.76  \\
                          & Noise         &  37.86  &  34.90  &   26.64   &  24.66  \\
                          & Ours     &  37.62  &  34.18  &   \textbf{32.72}   &   \textbf{29.56} \\ \hline
\multirow{3}{*}{ModeRec$\uparrow$}  & Deterministic &  29.99  &  22.51  &   27.42   &  23.37  \\
                          & Noise  &  34.24  &  29.98  &   36.50   &   31.43 \\
                          & Ours   &  \textbf{36.13}  &  \textbf{34.25}  &    \textbf{37.62}  &  \textbf{34.88}  \\ \hline
\end{tabular}}
\label{table:probabilistic2}
\end{table}

\begin{table*}[t]
\centering
\caption{Evaluation results of probabilistic modeling metrics on CrossTask. Joint training with horizon concatenation is applied to the Unet based model.}
\resizebox*{0.7\linewidth}{!}{
\renewcommand\arraystretch{1}
\begin{tabular}{cccccccccc}
\hline
& & \multicolumn{4}{c}{no task} & \multicolumn{4}{c}{task concatenation} \\ \cline{3-10}
Metric & Model  & $T$ = 3         & $T$ = 4           & $T$ = 5           & $T$ = 6    & $T$ = 3         & $T$ = 4           & $T$ = 5           & $T$ = 6           \\ \hline
\multirow{3}{*}{NLL$\downarrow$}  & Deterministic &  3.62 &  4.23   &   4.74   &  5.11  & 3.62  &   4.17  &  4.61  &  5.03  \\
                     & Noise         &  3.52 &  4.06   &   4.43   &  4.78  &  3.52 &  4.07   & 4.45  & 4.80 \\
                     & Ours          & \textbf{2.96}  &  \textbf{3.26}   &   \textbf{3.49}   &  \textbf{3.58}  & \textbf{2.93}  &  \textbf{3.19}   &   \textbf{3.40}   &  \textbf{3.48} \\ \hline
\multirow{3}{*}{KL-Div$\downarrow$} & Deterministic & 3.04  &   3.35  &   3.59   &  3.80  & 3.04  &  3.29   & 3.46 & 3.72 \\
                     & Noise   & 2.94  &  3.17   &   3.28   &  3.48  &  2.94 &  3.19   & 3.29  &   3.49       \\
                     & Ours   &  \textbf{2.38} &  \textbf{2.37}   &   \textbf{2.33}   &  \textbf{2.27}  &  \textbf{2.35} &  \textbf{2.30}   &   \textbf{2.24}   &  \textbf{2.17} \\ \hline
\multirow{3}{*}{SR$\uparrow$}       & Deterministic & 36.54 & 22.08     & 12.04   & 8.05  & 37.37  &   22.75  & 13.29 &   7.83      \\
                          & Noise         & 36.44          & 21.89          & 13.14          & 7.96     & 36.72  &  21.48   & 13.50 &   7.92   \\
                          & Ours          & \textbf{37.96}          & \textbf{22.56} & \textbf{14.30} & \textbf{8.93} & \textbf{37.46}  &   \textbf{22.83}  & \textbf{13.79} & \textbf{8.65}\\ \hline
\multirow{3}{*}{ModePrec$\uparrow$} & Deterministic &  54.64 &   46.14  &   33.56   &  21.57  & \textbf{55.27}  &   \textbf{47.48}  & \textbf{36.80} & 24.96 \\
                          & Noise         &  53.77 &   45.55  &   34.73   &  24.21  & 54.06 &  45.27   & 35.41 & 24.08 \\
                          & Ours     &  \textbf{55.31} &  \textbf{47.38}   &   \textbf{36.79}   &  \textbf{25.94}  &   54.86 &  46.96   &   36.75   &  \textbf{25.48} \\ \hline
\multirow{3}{*}{ModeRec$\uparrow$}  & Deterministic & 33.15  &  18.56   &   10.12   &  6.00  & 33.67  &   19.54  & 11.39 & 6.59 \\
                          & Noise  &  37.34 &   23.41  &   14.69   &  9.62  &  38.13 &  23.47   & 15.04 &  9.83  \\
                          & Ours   &  \textbf{52.44} &  \textbf{43.99}   &   \textbf{36.90}  &  \textbf{31.59}  &  \textbf{53.77} &  \textbf{44.90}   &   \textbf{38.68}  &  \textbf{33.44} \\ \hline
\end{tabular}}
\label{table:probabilistic1}
\end{table*}

\subsection{\re{Discussion on Computational Efficiency}}
\label{exp:efficiency}
\re{The computational efficiency of PDPP is provided in \cref{table:efficiency}, including the number of training parameters, FLOPs and time required for a single sampling of PDPP across different datasets. We have also studied the effect of joint training on model parameter savings and training cost. Experimental results demonstrate that joint training effectively saves model training parameters and training steps when planning step $T > 1$, which also enables flexible multi-step procedure predicting.  All the experimental results are obtained on a single  NVIDIA TITAN XP GPU.}

\subsection{Evaluation on probabilistic modeling}
\label{exp:uncertainty}
As discussed in \cref{sec::intro}, we introduce diffusion model to procedure planning to model the uncertainty in this problem. Here we follow\cite{DBLP:conf/cvpr/0004HDDWJ22} to evaluate our probabilistic modeling.

Our model is probabilistic by starting from random noises and denoising step by step. We here introduce two baselines to compare with our diffusion based approach. We first remove the diffusion process in our method to establish the \textit{Noise} baseline, which just samples from a random noise with the given observations and task class condition in one shot. Then we further establish the \textit{Deterministic} baseline by setting the start distribution $\hat{x}_N = 0$, thus the model directly predicts a certain result with the given conditions. For the \textit{Deterministic} baseline, we just sample once to get the plan since the result for \textit{Deterministic} is certain when observations and task class conditions are given. For the \textit{Noise} baseline and our diffusion based model, we sample 1500 action sequences as our probabilistic result to calculate the uncertain metrics. Note that the multiple sampling process is only required while evaluating probabilistic modeling and our model can generate a good plan just by sampling once.

We reproduce the \textit{KL divergence, NLL, ModeRec} and \textit{ModePrec} in\cite{DBLP:conf/cvpr/0004HDDWJ22} and use these metrics along with \textit{SR} to evaluate our probabilistic model. Specifically, the \textit{SR} and \textit{ModePrec} can reflect the accuracy of planning results, \textit{ModeRec} reflects the diversity of plans, and \textit{KL divergence, NLL} indicate the overall agreement between predictions and the ground truth distribution.

~\\ \noindent \textbf{Role of diffusion process.} Results in \cref{table:probabilistic1,table:probabilistic2} suggest that PDPP achieves the best performance on most metrics, which shows our approach has an excellent ability to model the uncertainty in procedure planning and can produce both

\begin{table*}[t]
\centering
\caption{Study on influence of joint training and DDIM sampling for probabilistic modeling results with PDPP.}
\resizebox*{0.7\linewidth}{!}{
\renewcommand\arraystretch{1}
\begin{tabular}{cccccccccc}
\hline
& & \multicolumn{4}{c}{CrossTask} & \multicolumn{2}{c}{COIN} & \multicolumn{2}{c}{NIV} \\ \cline{3-10}
Metric & Train\&Sample  & $T$ = 3         & $T$ = 4           & $T$ = 5           & $T$ = 6    & $T$ = 3         & $T$ = 4           & $T$ = 3           & $T$ = 4           \\ \hline
\multirow{3}{*}{NLL$\downarrow$}  & Separate\_DDIM &  3.61  &   3.85   &   3.77   &   4.06  &  5.15  &  5.47   &  4.93  &   4.75 \\
                     & Join\_DDIM         &  \textbf{2.93}  &   \textbf{3.19}   &    \textbf{3.40}   &  \textbf{3.48}   &  \textbf{5.03}  &  \textbf{5.20}   &  \textbf{4.77}  &  \textbf{4.58} \\
                     & Join\_DDPM &  3.14   &  3.42  &  3.65  &  3.72  &  5.13  &  5.41  &  5.01   &  4.95 \\
                      \hline
\multirow{3}{*}{KL-Div$\downarrow$} & Separate\_DDIM &  3.03  &  2.96    &    2.62   &   2.76  &  4.20  &  4.07   &  4.85  &  4.62 \\
                     & Join\_DDIM   & \textbf{2.35}  &  \textbf{2.30}   &   \textbf{2.24}   &  \textbf{2.17}  &  \textbf{4.08} &   \textbf{3.79}   &  \textbf{4.69} &    \textbf{4.46}  \\
                     & Join\_DDPM &  2.56   &   2.54  &  2.49  &  2.42  &  4.18  &  4.00   &   4.92  &  4.82 \\
                     \hline
\multirow{3}{*}{SR$\uparrow$}       & Separate\_DDIM &  37.20  &   21.48   &   13.45    &   8.41  &  \textbf{31.05}  &  22.45   &  30.20  &  26.67 \\
                          & Join\_DDIM         &  37.46  &   22.83   &   13.79    &  8.65   &  30.12  &  22.24   & 30.74   &  27.78 \\
                          & Join\_DDPM &  \textbf{38.07}   &  \textbf{23.79}   &  \textbf{14.75}  &  \textbf{9.52}  &  30.34  &   \textbf{22.68}  &  \textbf{30.85}   &  \textbf{27.96} \\
                          \hline
\multirow{3}{*}{ModePrec$\uparrow$} & Separate\_DDIM &  53.14  &   44.55   &   36.30    &  25.61   &  \textbf{38.21}  &  33.66   &  31.78  & 29.10 \\
                          & Join\_DDIM       &  54.86  &   46.96   &    36.75   &  25.48   &  37.62  &  34.18   &  32.72  & 29.56 \\
                          & Join\_DDPM &  \textbf{55.12}   &   \textbf{48.02}  &  \textbf{38.44}  &  \textbf{27.78}  &  37.81  &  \textbf{34.68}   &  \textbf{33.26}   & \textbf{30.49} \\
                          \hline
\multirow{3}{*}{ModeRec$\uparrow$}  & Separate\_DDIM &  36.49  &   31.10   &  29.45     &  22.68   &  34.63  &  29.64   &  33.09  & 33.08 \\
                          & Join\_DDIM &  \textbf{53.77}  &  \textbf{44.90}    &   \textbf{38.68}    &   \textbf{33.44}  &  \textbf{36.13}  &  \textbf{34.25}   &   \textbf{37.62} & \textbf{34.88} \\
                          & Join\_DDPM &  45.96   &  36.66   &  30.97  &  26.50  &  33.90  &   28.96  &  34.33   & 32.66 \\
                          \hline
\end{tabular}}
\label{table:probabilistic_joint}
\end{table*}

\noindent diverse and reasonable plans. Specifically, \textit{Deterministic} baseline generates more accurate results while \textit{Noise} baseline gives more diverse results when $T=3,4$. Longer horizons bring more diverse plans, thus the \textit{Noise} baseline can achieve better SR results when $T=5,6$ since the added noise helps to model the uncertainty. Our diffusion based PDPP makes a good trade-off between the diversity and accuracy of planning, thus generates both accurate and diverse results that best match the ground truth.

~\\ \noindent \textbf{Influence of joint training and DDIM sampling.} In this section, we study how joint training and DDIM sampling can influence the modeling of uncertainty. For NIV and COIN, we just apply the best setting as described in the end of \cref{exp:ablation}. For CrossTask, we here use the Unet based model trained with task and horizon concatenation instead since its $NLL$ and $KL-Div$ metrics are better than training without task condition(shown in \cref{table:probabilistic1}). The results are shown in \cref{table:probabilistic_joint}. Results of $Separate\_DDIM$ and $Join\_DDIM$ indicate that joint training can further improve the \textit{ModeRec} performance and generates more diverse plans. So we can draw the conclusion that joint training has a positive effect on modeling the uncertainty. As for DDIM sampling, we can know that sampling with DDPM gets more precise results and DDIM sampling helps for planning diversely. One explanation for this is that the DDPM sampling is consistent with our training process, while DDIM sampling conducts skip sampling for acceleration. Thus DDIM sampling can deviate from the intended sampling route slightly and brings more diversity to planning results.

\begin{figure*}[t]
  \centering
  \includegraphics[width=1\linewidth]{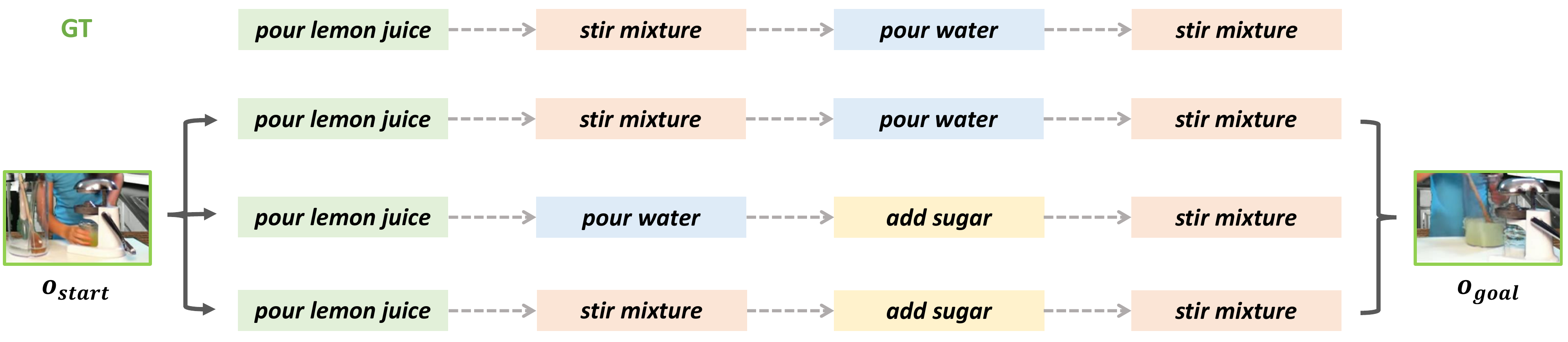} 
   \caption{Visualization of diverse plans produced by our model with horizon $T = 4$ on CrossTask. }
   \label{fig:sup2}
\end{figure*}

\begin{figure*}[t]
  \centering
  \includegraphics[width=1\linewidth]{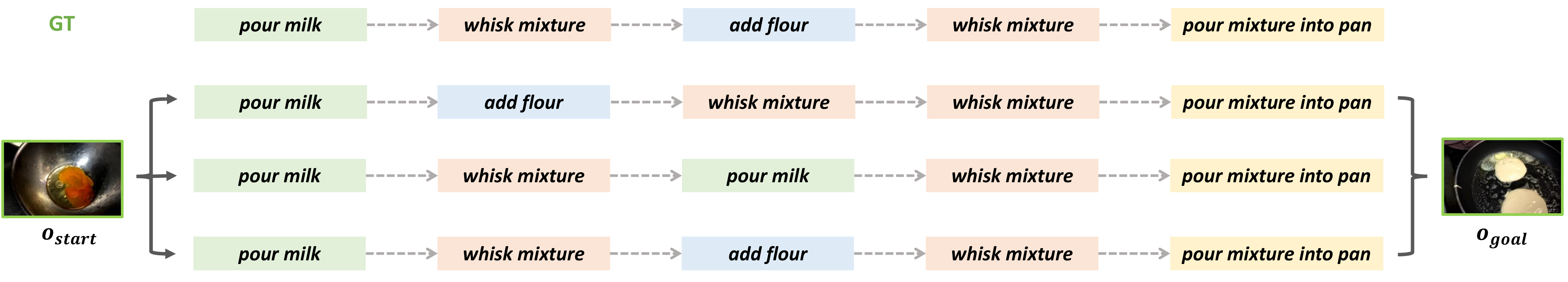}
   \caption{Visualization of diverse plans produced by our model with horizon $T = 5$ on CrossTask.}
   \label{fig:sup3}
\end{figure*}

~\\ \noindent \textbf{Visualization Results.}
In \cref{fig:sup2,fig:sup3}, we show the visualizations of different plans with the same start and goal observations produced by our PDPP for different prediction horizons on CrossTask. In each figure, the images denote the start and goal observations, the first row denotes the ground truth actions (rows with "GT") and other rows denote multiple reasonable plans produced by our model, respectively. Here the reasonable plans are plans that share the same start and end actions with the ground truth plan and exist in the test dataset. We can see that our model can generate both right and diverse plans with the given observations.

\begin{table}[]
\centering
\caption{Evaluation results on CrossTask and COIN for VPA with prediction horizon $T \in \{3, 4\}$. Here "Random\_g" denotes the $Random$ $w/goal$ baseline, "P, p1, p2" means Protocol, protocol1 and protocol2, respectively.}
\resizebox*{1.0\linewidth}{!}{
\begin{tabular}{lcccccccc}
\hline
       &          &             & \multicolumn{3}{c}{$T$ = 3} & \multicolumn{3}{c}{$T$ = 4} \\ \cline{4-9} 
\multicolumn{1}{l}{Dataset}    & \multicolumn{1}{l}{Model}       & P & SR$\uparrow$    & mAcc$\uparrow$   & mIoU$\uparrow$   & SR$\uparrow$    & mAcc$\uparrow$   & mIoU$\uparrow$   \\ \hline
\multirow{5}{*}{Crosstask} & \multicolumn{1}{l}{Random}       &     p1        &   0.0     &   0.9   &    1.5    &   0.0     &   0.9    &    1.9     \\
& \multicolumn{1}{l}{Random\_g}       &     p1        &    0.3    &   13.4   &    23.6    &    0.0    &   12.7    &   27.8   \\
& \multicolumn{1}{l}{DDN\cite{DBLP:conf/eccv/ChangHXAFN20}}       &     p1       &   6.8     &   25.8   &    35.2    &     3.6   &   24.1    &  37.0  \\
& \multicolumn{1}{l}{VLaMP\cite{patel2023pretrained}}       &     p1        &   10.3     &   35.3   &   44.0     &    4.4    &   31.7    &  43.4  \\
& \multicolumn{1}{l}{PDPP}       &     p1       &    \textbf{17.5}    &   \textbf{48.5}   &   \textbf{55.3}     &   \textbf{9.8}     &   \textbf{44.3}    & \textbf{56.6}  \\
& \multicolumn{1}{l}{PDPP}       &     p2       &   11.6     &    36.7  &    47.7    &    6.3    &    35.1   & 50.9 \\
\hline
\multirow{5}{*}{COIN} & \multicolumn{1}{l}{Random}       &     p1 &    0.0   &    0.1    &    0.2    &   0.0    &  0.1 & 0.2 \\
& \multicolumn{1}{l}{Random\_g}       &     p1 &    1.7    &   21.4   &    42.7    &   0.3     &   20.1    &  47.7 \\
& \multicolumn{1}{l}{DDN\cite{DBLP:conf/eccv/ChangHXAFN20}}       &     p1 &   10.1    &   22.3   &   32.2     &    7.0    &    21.0   &   37.3   \\
& \multicolumn{1}{l}{VLaMP\cite{patel2023pretrained}}       &     p1   &     18.3   &   39.2   &   56.6     &   9.0   &   35.2    &  54.2   \\
& \multicolumn{1}{l}{PDPP}       &     p1     &    \textbf{25.5}    &   \textbf{52.9}   &   \textbf{67.3}     &    \textbf{19.2}    &    \textbf{51.0}   &  \textbf{71.5}  \\
& \multicolumn{1}{l}{PDPP}       &     p2    &   21.6     &   47.2   &    65.1    &    16.2    &   46.9    &  69.5 \\
\hline
\end{tabular}
}
\label{table:VPAexp}
\end{table}

\subsection{Results on the VPA task}
\label{exp:VPA}
We in this section apply our PDPP to the \revise{Visual Planners for human Assistance (VPA)} problem by removing the goal observation and using the ground truth task label as condition instead. \re{The input of VPA task includes the current observation, the goal description, and the video history. The video history is an additional input (not required by procedure planning) that records the entire process from the beginning to the current state. Although it can provide some contextual information, it is computationally expensive and difficult to achieve real-time planning in the existing frameworks. Therefore, we choose to only use the current observation as starting point to predict action sequence to reach the goal, which is more convenient and efficient.} We follow Patel \textit{et al.}\cite{patel2023pretrained} to evaluate our model on CrossTask and COIN datasets. For CrossTask, we use the Unet based model and add both the task and horizon conditions by concatenation. For COIN, we apply the Unet-attention model with task mask and horizon concatenation.

Take the start time of the first action to be predicted as $t_{start}$, Patel \textit{et al.}\cite{patel2023pretrained} gets their start observation $o_s$ by capturing the video clip from time $t_{start}-2$ to $t_{start}+2$, which we denote as "protocol 1". However, we here argue that since the aim for VPA problem is to assist human in everyday lives, taking video clip from time $t_{start}$ to $t_{start}+2$ is not suitable because actions after $t_{start}$ are inaccessible and have not been performed. Thus we select 4 seconds-long video content before $t_{start}$ as $o_s$, which we denote as "protocol 2".

We compare our model with all approaches in\cite{patel2023pretrained}: 

\noindent - $Random$. This policy randomly selects actions from the available action space in dataset to produce the plans.

\noindent - $Random$ $w/goal$. This policy randomly selects actions from the task-limited action space in dataset to produce the plans. Only actions applicable to the given task class can be involved.

\noindent - $DDN$\cite{DBLP:conf/eccv/ChangHXAFN20}. DDN model is a two-branch autoregressive model, which learns an abstract representation of action steps and tries to predict the state-action transition in the feature space.

\noindent - $VLaMP$\cite{patel2023pretrained}. VLaMP implements their planning model with pretrained transformer-based language model and predicts action and observation tokens autoregressively. Beam search is applied for inference. Besides, video history is required and processed by the segmentation module for better planning.

\cref{table:VPAexp} presents the experiment results for VPA. Our PDPP achieves the state-of-the-art performance, even without the video history. This again shows the great generalization ability of our PDPP on goal-directed planning in instructional videos and the effectiveness of modeling action sequence as a whole.

\section{Conclusion}

In this paper, we have casted procedure planning in instructional videos as a distribution fitting problem and addressed it with a projected diffusion model PDPP. Compared with previous work, our model requires less supervision and can be trained with a simple learning scheme. PDPP performs diffusion process both for training and sampling, thus models the uncertainty well and can generate reasonable and diverse plans.  We
instantiate our PDPP with three popular architectures for diffusion model and conduct extensive ablation study on the design and condition introduction of our model. We also apply joint training to PDPP to plan with multiple horizons and save training parameters.
Evaluation results on three datasets with different scales demonstrate that our model obtains a notable improvement over previous approaches among multiple planning horizons. Our work demonstrates that modeling action sequence as a whole distribution is an effective solution to goal-directed planning in instructional videos. In the future, we consider extending PDPP to open domain planning with language described action and task labels.

\section*{Acknowledgments}
This work is supported by the National Key R$\&$D Program of China (No. 2022ZD0160900), the National Natural Science Foundation of China (No. 62076119), the Fundamental Research Funds for the Central Universities (No. 020214380119), Jiangsu Frontier Technology Research and Development Program (No. BF2024076), and the Collaborative Innovation Center of Novel Software Technology and Industrialization.

{\small

\bibliographystyle{ieee_fullname}
\bibliography{ref}
}

  

\end{document}